\documentclass{article}

\usepackage[a4paper, total={6in, 9in}]{geometry}
\usepackage[square,sort,comma,numbers]{natbib}
\usepackage[title]
  {appendix}
\usepackage[utf8]{inputenc} 
\usepackage[T1]{fontenc}    
\usepackage{hyperref}       
\usepackage{url}            
\usepackage{booktabs}       
\usepackage{amsfonts}       
\usepackage{nicefrac}       
\usepackage{microtype}      
\usepackage{url}            
\hypersetup{colorlinks, citecolor=blue, linkcolor=red, urlcolor=blue}
\usepackage{mathtools}
\usepackage[noblocks]{authblk}
\usepackage{graphicx}

\def\equationautorefname~#1\null{(#1)\null}

\usepackage{graphicx}
\usepackage{epstopdf}
\usepackage{amsopn}

\usepackage[bb=boondox]{mathalfa}
\usepackage{comment}
\usepackage{algorithm}
\usepackage{algpseudocode}
\algdef{SE}[REPEATN]{RepeatN}{End}[1]{\algorithmicrepeat\ #1 \textbf{times}}{\algorithmicend}

\ifpdf
  \DeclareGraphicsExtensions{.eps,.pdf,.png,.jpg}
\else
  \DeclareGraphicsExtensions{.eps}
\fi

\usepackage[capitalize,noabbrev]{cleveref}

\usepackage{enumitem}
\setlist[enumerate]{leftmargin=.5in}
\setlist[itemize]{leftmargin=.5in}

\newtheorem{definition}{Definition}

\newtheorem{theorem}{Theorem}

\usepackage{float} 
\usepackage{xr}
\usepackage{wrapfig}
\usepackage[font=small,labelfont=bf]{caption}
\usepackage{subcaption}
\usepackage{tikz}
\usepackage{accents}
\usepackage{array}
\usepackage{tabularx, makecell, booktabs}

\usepackage{xcolor}
\usepackage{pifont}

\usepackage{makecell}

\usepackage{selectp}
\usepackage{bm} 
\usepackage{amsmath}
\usepackage{amssymb}
\usepackage{mathtools}

\usepackage{defs}
\PassOptionsToPackage{usedvipsnames}{xcolor}
\definecolor{custom_blue}{rgb}{0.21,0.49,0.74}
\PassOptionsToPackage{pagebackref,breaklinks,colorlinks,allcolors=custom_blue}{hyperref}

\title{Disentangling Safe and Unsafe Corruptions via ~\\ Anisotropy and Locality}

\author[1]{Ramchandran Muthukumar\thanks{Corresponding author: \tt{rmuthuk1@jhu.edu}}}
\author[2]{Ambar Pal\footnote{This work is not related to AP's position at Amazon}}
\author[1]{Jeremias Sulam}
\author[3]{Ren\'e Vidal}

\affil[1]{Johns Hopkins University}
\affil[2]{Amazon Web Services}
\affil[3]{University of Pennsylvania}

\begin{document}
\maketitle

\begin{abstract}
State-of-the-art machine learning systems are vulnerable to small perturbations to their input, where ``{small}'' is defined according to a threat model that assigns a positive threat to each perturbation. Most prior works define a task-agnostic, isotropic, and global threat, like the $\ell_p$ norm, where the magnitude of the perturbation fully determines the degree of the threat and neither the direction of the attack nor its position in space matter. However, common corruptions in computer vision, such as blur, compression, or occlusions, are not well captured by such threat models. This paper proposes a novel threat model called \texttt{Projected Displacement} (PD) to study robustness beyond existing isotropic and global threat models. The proposed threat model measures the threat of a perturbation via its alignment with \textit{unsafe directions}, defined as directions in the input space along which a perturbation of sufficient magnitude changes the ground truth class label. Unsafe directions are identified locally for each input based on observed training data. In this way, the PD threat model exhibits anisotropy and locality. The PD threat model is computationally efficient and can be easily integrated into existing robustness pipelines. Experiments on Imagenet-1k data indicate that, for any input, the set of perturbations with small PD threat includes \textit{safe} perturbations of large $\ell_p$ norm that preserve the true label, such as noise, blur and compression, while simultaneously excluding \textit{unsafe} perturbations that alter the true label. Unlike perceptual threat models based on embeddings of large-vision models, the PD threat model can be readily computed for arbitrary classification tasks without pre-training or finetuning. Further additional task information such as sensitivity to image regions or concept hierarchies can be easily integrated into the assessment of threat and thus the PD threat model presents practitioners with a flexible, task-driven threat specification that alleviates the limitations of $\ell_p$-threat models.

\end{abstract}

\section{Introduction}
\label{sec: intro}
Modern machine learning (ML) systems are nearing widespread deployment in civilian life, and hence a comprehensive understanding of their security vulnerabilities is necessary \cite{1224151}. One such vulnerability concerns the ability of a malicious adversary to tamper with predictions by adding small imperceptible corruptions referred to as \emph{adversarial perturbations} \citep{Szegedy2014IntriguingPO, Goodfellow2015ExplainingAH}. There is by now overwhelming evidence that carefully crafted adversarial perturbations can foil the prediction of state-of-the-art machine learning classifiers, i.e., they are not adversarially robust \citep{papernot2016limitations, Carlini2017AdversarialEA, croce2021robustbench}. Such vulnerabilities have been observed in a wide variety of applications like computer vision \citep{wu2020makinginvisibilitycloakreal, akhtar2021advances}, speech recognition \citep{qin2019imperceptible, żelasko2021adversarialattacksdefensesspeech}, autonomous driving \citep{Cao2019AdversarialSA, pmlr-v164-tu22a}, and more. The goal of designing safe and reliable ML systems remains incomplete \citep{debenedetti2024scaling, pmlr-v235-bartoldson24a} despite significant investment \citep{gowal2021uncoveringlimitsadversarialtraining, gowal2021improvingrobustnessusinggenerated,liu2023comprehensivestudyrobustnessimage}. Often, strategies that intend to foil performance, i.e., \textit{adversarial attacks} \citep{Goodfellow2015ExplainingAH, Kurakin2017AdversarialEI, Carlini2017AdversarialEA, Tramr2020OnAA, 10.5555/3524938.3525144}, have proven more successful than strategies aimed at mitigating vulnerabilities, i.e., \textit{adversarial defenses} \citep{Madry2018TowardsDL, pmlr-v80-wong18a, pmlr-v97-zhang19p, wang2023betterdiffusionmodelsimprove,cui2023decoupledkullbackleiblerdivergenceloss, bai2024improvingaccuracyrobustnesstradeoffclassifiers}. 

Meaningfully evaluating the adversarial robustness of a machine learning system requires a formal specification of a \textit{threat function}, and an associated \textit{threat model} that limits the scope of malicious adversaries \citep{gilmer2018motivatingrulesgameadversarial}. Informally, a threat function $d: \mathbb{R}^d \times \mathbb{R}^d \rightarrow \mathbb{R}^{\geq 0}$ measures the threat\footnote{The flexibility to define threat w.r.t. input $\x$ is critical to our work.} represented by a corruption $\bdel \in \mathbb{R}^d$ towards altering the true label at input $\x \in \mathbb{R}^d$ as $d(\x,\bdel)$. We let $\mathcal{S}(\x, d, \varepsilon) \coloneqq \{\bdel \in \mathbb{R}^d ~|~ d(\x,\bdel) \leq  \varepsilon\}$ denote the $\varepsilon$-sublevel set of perturbations where the threat $d$ is measured w.r.t.~input $\x$. A threat model $(d,\varepsilon)$ is a pair of a threat function $d$ and a permissible threshold $\varepsilon$ that together define the set of permissible perturbations $\mathcal{S}(\x, d, \varepsilon)$. Under the threat model $(d,\varepsilon)$, the \textit{robust accuracy} of a classifier $h \in \cH$ is the probability that the label prediction at $\x$ is locally invariant to corruptions within the permissible set $\mathcal{S}(\x, d, \varepsilon)$; see \Cref{def:TM-RA} for an explicit definition.

One of the most commonly used threat models is the $\ell_p$-threat model $(d_p, \epsilon)$, corresponding to the choice\footnote{For any vector $\bdel \in \mathbb{R}^d$, the $\ell_p$ norm $\norm{\cdot}_p$ for $p \geq 1$ is defined as $\norm{\bdel}_p\coloneqq \left(\sum_{i} |\delta_i|^p \right)^{\frac{1}{p}}$.} $d_p(\x, \bdel) \coloneqq \norm{\bdel}_p$. The threat function $d_p$ is task-agnostic, easy to evaluate, and induces a compact sub-level set $\mathcal{S}(\x, d_p, \varepsilon)$ that allows for efficient projection. Hence, the $\ell_p$-threat model $(d_p, \varepsilon)$ presents a natural starting point for investigating robustness. RobustBench \citep{croce2021robustbench} maintains an up-to-date leaderboard of the robust accuracies of benchmark models under $(d_p, \varepsilon)$. Unfortunately, the progress towards achieving perfect adversarial robustness (i.e., 100$\%$ robust test accuracy) in RobustBench has plateaued in recent years. Unlike supervised learning on clean data, scaling data, model size and computing resources might be insufficient to bridge the gap \citep{debenedetti2024scaling, pmlr-v235-bartoldson24a}.  Below we expand further on the fundamental limitations of isotropic and global threat models.

\subsection{Motivation: Specification of Threat Model}
\label{subsec: motivation}
For image-based datasets, it is widely recognized that $\ell_p$ norms are neither necessary nor sufficient for capturing perceptual similarity \cite{Sharif2018OnTS, 10.1007/978-3-030-64793-3_10}, which poses significant challenges for accurately evaluating robustness \citep{jacobsen2018excessive, DBLP:conf/icml/TramerBCPJ20}. 
\begin{figure}[h]
\centering
\includegraphics[width=0.5\linewidth]{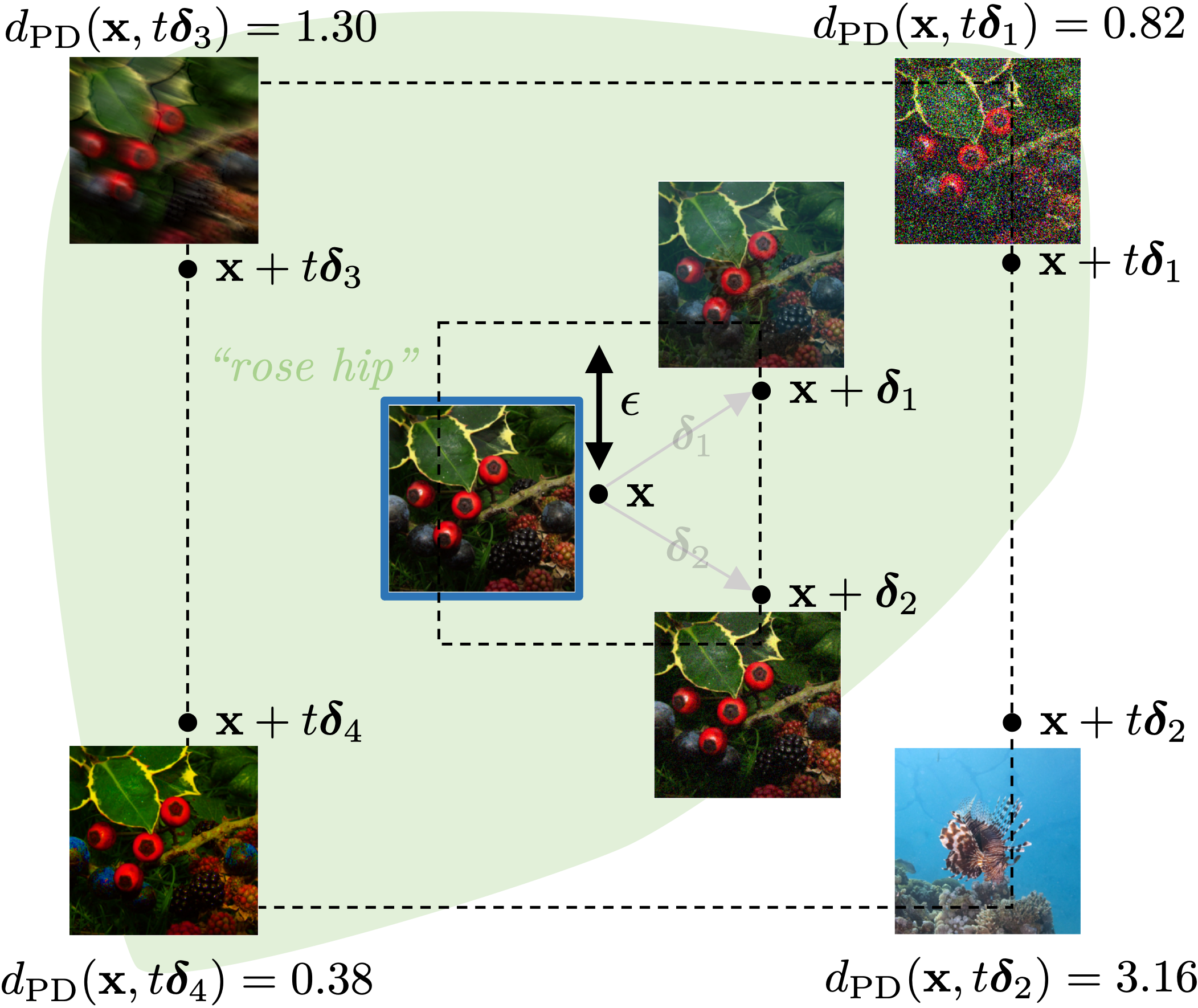}
\caption{Corruptions with equal $\ell_{\infty}$-threat, $\norm{\bdel_1}_\infty = \norm{\bdel_2}_\infty = \norm{\bdel_3}_\infty = \norm{\bdel_4}_\infty$, but varying PD threat.}
\label{fig:imagenet-motivation-1}
\end{figure}
In \Cref{fig:imagenet-motivation-1}, $\x$ (at the center) is an image of class \textsc{rose hip} from the Imagenet-1k dataset, and $\x + \bdel_1$, $\x + \bdel_2$ are two corrupted images equidistant (w.r.t. $\ell_{\infty}$) from $\x$, i.e., $\norm{\bdel_1}_{\infty} = \norm{\bdel_2}_{\infty} = \varepsilon$. Intuitively, both $\x+\bdel_1$ and $\x+\bdel_2$ share the same label as $\x$, since its appearance still depicts the fruit \textsc{rose hip}. We refer to such perturbations that preserve the class label as \textit{safe}, while those that change it are termed \textit{unsafe}. Consider now moving further along these directions to $\x + t\bdel_1$ and $\x + t\bdel_2$, for $t>1$. These perturbed points are again equidistant from $\x$, but while $t\bdel_1$ does not alter the true label (i.e., it is safe), $t\bdel_2$ does (and it is unsafe). As a result, any threat model $(d_{\infty}, t' \varepsilon)$ with $t' \geq t$ necessarily incurs misspecification: any classifier $h$ stable to all perturbations in $\mathcal{S}(\x, d_{\infty},t' \varepsilon)$ produces an incorrect prediction at $\x+t\bdel_2$, i.e., $h(\x + t\bdel_2) = \textsc{rose hip}$. On the other hand, shrinking the threat model is also futile, as then the safe perturbation $t\bdel_1$ is necessarily excluded. 

This illustrates that standard $\ell_p$-threat models are unable to distinguish between the safe perturbations that preserve the true label and the unsafe perturbations that alter the true label. Hence, perfect robust accuracy under $\ell_p$ threat models might be neither achievable nor desirable. Attempting to resolve these limitations, \citet{laidlaw2021perceptualadversarialrobustnessdefense, Luo_2022_CVPR, Chen_2023_ICCV}  formulate perceptual threat models via neural \textit{perceptual}  metric based on neural representations. Unfortunately (but unsurprisingly), all neural perceptual distance metrics are themselves vulnerable to adversarial attacks \cite{kettunen2019elpipsrobustperceptualimage, Ghildyal2023AttackingPS, ghazanfari2023rlpips, Ghazanfari2023LipSimAP}.%

\subsection{Summary of Contributions}
\label{subsec: contrib-summary}
We propose a novel threat function called \textit{Projected Displacement} (PD),  denoted by $d_{\mathrm{PD}}$ (see \Cref{def:noniso-threat-k} for an explicit description), that accounts for the local directional statistics of the data via unsafe directions\footnote{\textit{Directions} refer to vectors $\vc{u} \in \mathbb{R}^d$ with unit $\ell_2$ norm.} $\mathcal{U}(\x)$ (\Cref{def:unsafe-observed-k}) at each input $\x$. For an input $\x$ with label ${y}$, any input $\xtil$ with label $\tilde{y} \neq y$ represents the unsafe direction $\vc{u} \coloneqq \frac{\xtil-\x}{\norm{\xtil-\x}_2}$, which will be estimated from observed training data. Then, the degree of threat of each perturbation $\bdel$ is defined as the maximal ``alignment'' with the set of unsafe directions: A large threat value $d_{\mathrm{PD}}(\x, \bdel)$ indicates that $\bdel$ is well-aligned with an unsafe direction $\vc{u} \in \mathcal{U}(\x)$. We highlight key properties of the PD threat model.
\begin{enumerate}
    \item \textbf{Disentangling safe vs unsafe perturbations}.
    PD threat is able to distinguish between safe and unsafe perturbations of equal $\ell_p$ norms. From \Cref{fig:imagenet-motivation-1}, different corruptions (Gaussian noise $\bdel_1$, motion blur $\bdel_3$ and saturation $\bdel_4$) applied to the original image $\x$ all have equal $\ell_\infty$ norm of $240/255$. Despite this, the true label of these perturbed samples does not change, unlike that of corruption $\bdel_2$, which results in an image $\x+t\bdel_2$ with a different class. For each perturbation $t\bdel_{i}$, the computed values of PD threat in \Cref{fig:imagenet-motivation-1} naturally reflect how close they are to changing the class. \Cref{Sec:Experiments} presents experimental evidence that PD threat is competent with a state-of-the-art neural perceptual threat model DreamSim \cite{fu2024dreamsim} on distinguishing safe and unsafe perturbations. 

    \item \textbf{Ease of use}. 
    As we will show, the sub-level set $\mathcal{S}(\x, d_{\rm PD},\varepsilon)$ at each input $\x$ is convex, allowing for efficient projections onto it. This enables a straightforward plug-and-play mechanism for adapting existing adversarial attacks to the PD threat model. We incorporate such an extension of \textit{AutoAttack} \citep{10.5555/3524938.3525144} to evaluate the robustness under the PD threat model $(d_{\rm PD}, \varepsilon)$ for state-of-the-art robust classifiers registered at RobustBench \citep{croce2021robustbench}. 

    \item \textbf{Task-dependency.}
    PD threat models can readily integrate additional task-relevant information via label annotation, such as inter-class hierarchy (e.g., the Wordnet hierarchy) or pixel annotation (e.g., segmentation masks)--see \Cref{tab:flexibile-threat-model}-- allowing for variations in the form of robustness while retaining the above benefits (anisotropy, locality, convex sub-level sets, and efficient projection). %

    \begin{table}[h]
    \centering
    \resizebox{\columnwidth}{!}{%
    \begin{tabular}{|c|c|c|c|c|}
    \toprule
    \textsc{Property} &  \textsc{Features} & Norm-Based & DreamSim \cite{fu2024dreamsim} & Projected Displacement \\
    \midrule
Degree of Threat
     & Anisotropy, Locality & \redcross & \greencheck & \greencheck \\ 
     \midrule
     Label Annotation & Inter-class Distance & \redcross & ${\color{orange} \sim}$ & \greencheck \\
     \midrule
     Pixel Annotation & Segmentation & \greencheck & \redcross & \greencheck\\
     \bottomrule
    \end{tabular}
    }
    \caption{Comparison of Threat Models}
    \label{tab:flexibile-threat-model}
    \end{table}
\end{enumerate}

\subsection{Related Work}
\label{subsec: relatedwork}
In this section, we highlight prior work that explores robustness beyond $\ell_p$-threat models. 

\textbf{Common Corruptions.}
Neural networks are vulnerable to image distortions (e.g. translation, rotation), common corruptions (e.g. blur, noise), changes in lighting, view, depth, obfuscations, etc. \cite{dodge2017study, NEURIPS2018_0937fb58, xiao2018spatially, 8578565, hendrycks2019robustness, NEURIPS2021_1d497805, kar20223d, stimberg2023benchmarking}. Note that these are \emph{safe perturbations} as they preserve the true label, even if they have large $\ell_p$ norms. To bridge this gap, \cite{hendrycks2019robustness} formalized the notion of \textit{corruption robustness} and presented a standardized benchmark dataset, \textbf{Imagenet-C},  containing 19 styles of common image corruptions (categorized into \textit{noise}, \textit{blur}, \textit{weather}, and \textit{digital}). Training on \textbf{Imagenet-C}, or on other augmentations, can result in over-fitting, particularly to specific types of corruptions \cite{NEURIPS2021_1d497805}. Additionally,  \cite{NEURIPS2021_1d497805} introduced a new dataset, \textbf{Imagenet-$\bar{\text{C}}$}, with 10 new corruption styles identified from a larger set of real-world corruptions. These will be useful later in our experiments. 

\textbf{Perceptual Distance Metrics.}
Since $\ell_p$ norms provide inaccurate approximations to perceptual distances, several works have attempted to propose alternatives, such as SSIM \citep{1284395}, FSIM \cite{5705575}, and HDR-VDP-2 \cite{Mantiuk2011HDRVDP2AC}. With the advent of learning data-driven representations, \cite{8578166, fu2024dreamsim} hail the effectiveness of internal representations of neural classifiers in capturing perceptual similarity. DreamSim \citep{fu2024dreamsim}, the current state-of-the-art among the neural perceptual distance metrics, is  fine-tuned on a large dataset of human perceptual similarity judgements.

\section{Key Definitions}
\label{sec: prelims}
\emph{Notation}. We denote scalar quantities in Roman letters and vectors in boldface Roman letters. The empty set is given by $\emptyset$. For any set $\mathcal{V}$, we denote by $2^{\mathcal{V}}$ the power set of all its subsets. For any set $\mathcal{V} \subset \mathbb{R}^d$, we denote the complement of the set w.r.t $\mathbb{R}^d$ by $\mathcal{V}^c$. For any set $\mathcal{V} \subset \mathbb{R}^d$ and a vector $\vc{w}\in \mathbb{R}^d$, $\mathcal{V}-\{\vc{w}\}$ is the Minkowski difference, i.e., the set $\{\vc{v} - \vc{w} ~|~ \forall\; \vc{v} \in \mathcal{V}\}$ (subtracting $\vc{w}$ from all elements of $\mathcal{V}$). We denote the unit ball and the unit sphere in $\mathbb{R}^d$ w.r.t. the $\ell_p$ norm by $\mathbb{B}^d_{p} \coloneqq \{\vc{v} \in \mathbb{R}^d ~|~ \norm{\vc{v}}_p\leq 1\}$ and $\mathbb{S}^d_p \coloneqq \{\vc{v} \in \mathbb{R}^d ~|~ \norm{\vc{v}}_p = 1\}$, respectively. A \textit{direction} is an element of the unit sphere $\mathbb{S}^d_2$. 

\subsection{Supervised Learning}
\label{subsec: supervised}
We consider a supervised learning task with a bounded input domain $\cX \subset \mathbb{R}^d$ and labels $\cY = \{1,\ldots, C\}$ such that each input $\x$ is assigned a label by a deterministic true\footnote{Realizability in the context of adversarial robustness has also been studied in \citet{DBLP:conf/icml/TramerBCPJ20, pmlr-v206-awasthi23c}.} labeling function $h^\star: \cX \rightarrow \cY$. The true labeling function $h^\star$ partitions the input domain $\cX \coloneqq \cup_{y} \cX_y$, where $\cX_{y}$ is the set of inputs assigned label $y$, i.e., $\cX_{y} = \{\x \in \cX\ | \ h^\star(\x) = y\}$. We assume that the partition sets $\cX_1, \ldots, \cX_C$ are nonempty and open in order to avoid degeneracies. 

We let $\cD_\cX$ be a data distribution over the input domain $\cX$, and $\cD_\cZ$ the extension\footnote{Sampling $\vc{z} \sim \cD_\cZ$ is equivalent to sampling $(\x, h^\star(\x))$ where $\x\sim \cD_\cX$.} of $\cD_\cX$ to a distribution over the joint domain $\cZ \coloneqq \cX \times \cY$ using the true labeling function $h^\star$. The true labeling function $h^\star$ and the marginal input distribution $\cD_\cX$ are unknown, but we observe finite labeled training data $\samp\coloneqq \{(\x^{(1)},y^{(1)}),\ldots,(\x^{(m)},y^{(m)})\} \overset{\mathrm{i.i.d.}}{\sim} (\cD_\cZ)^m$. For any classifier $h :\cX \rightarrow \cY$, its accuracy is the probability of correctly labeling a random input, i.e., $\mathrm{Acc}(h) \coloneqq \mathrm{Prob}_{\x \sim \cD_\cX} [h(\x) = h^\star(\x)]$. The task of supervised learning seeks a classifier $\bar{h}$ in a hypothesis class $\cH \subset \{h:\cX \rightarrow \cY \}$ with high accuracy.

\subsection{Adversarial Perturbations}
\label{subsec: adv-pert}
In this article, we formally define adversarial perturbations through the lens of the partition sets $\cX_y$. 
\begin{definition}[Adversarial Perturbation]\label{def:adv-rob}
An \emph{adversarial perturbation} $\bdel \in \mathbb{R}^d$ for predictor $h$ at input $\x$ is any corruption that is
\begin{enumerate}
    \item \textbf{Domain constrained:} $\x+\bdel \in \cX$, i.e., $\bdel \in \cX - \{\x\}$.
    \item \textbf{Label invariant:} $\x+\bdel$ has the same true label as $\x$, i.e., $h^\star(\x+\bdel) = h^\star(\x)$.
    \item \textbf{Adversarial:} $\x+\bdel$ is misclassified by $h$, i.e., $h(\x+\bdel) \neq h^\star(\x+\bdel)$.
\end{enumerate}
\end{definition}

This natural definition has the following important implications. First, the labels assigned by $h^\star$ are considered the ground truth and thus $h^\star$ has no adversarial perturbation at any input. The presence of adversarial perturbations indicates an imperfect labeling function $h$. Based on Definition \ref{def:adv-rob}, perturbations $\bdel \in \mathbb{R}^d$ such that $\x+\bdel \notin \cX$ are not deemed adversarial as the classifier $h$ is not required to predict labels outside the input domain. On the other hand, for $\x$ with true label $y$, a perturbation $\bdel$ such that $h^\star(\x+\bdel) \neq y$ cannot be adversarial at $\x$ as the corrupted input $\x+\bdel$ has a different \emph{true} label. Finally, if $h$ predicts the label at $\x$ incorrectly, then $\vc{0}$ is already an adversarial perturbation. 

\subsection{Adversarial Robustness via Threat Specification}
\label{subsec: adv-threat}
We investigate the adversarial robustness of machine learning classifiers by specifying a \textit{threat function} $d:\cX \times \mathbb{R}^d \rightarrow \mathbb{R}_{\geq 0}$. The $\varepsilon$-sublevel set of a threat function $d$ at an input $\x$, $\mathcal{S}(\x, d, \varepsilon)\coloneqq \{\bdel \in \mathbb{R}^d ~|~ d(\x,\bdel) \leq \varepsilon\}$, defines the set of perturbations with threat at most $\varepsilon$. Threat functions encode a way for a practitioner to specify the kind of robustness suitable for a particular learning task. Standard threat models based on $\ell_p$ norms correspond to the choice $d(\x,\bdel)\coloneqq\|\bdel\|_p$.  
\begin{definition}[Threat Model, Robust Accuracy]\label{def:TM-RA}
A threat model $(d,\varepsilon)$ is a pair of threat function $d$ and permissible threshold $\varepsilon$ that together define the set of permissible perturbations at each input $\x$ as the sublevel set $\mathcal{S}(\x, d, \varepsilon)$. 
Under the threat model $(d,\varepsilon)$, the \textit{robust accuracy} $\mathrm{RobustAcc}(h, (d,\varepsilon))$ of a classifier $h \in \cH$ is the probability of stable prediction upon corruptions within the permissible set, i.e., 
$
\mathrm{Prob}_{\x} \big[\forall\; \bdel \in \mathcal{S}(\x, d, \varepsilon), ~~h(\x+\bdel) = h^\star(\x)\big]. 
$
\end{definition}
Thus we define \textit{robustness} as the stability\footnote{Note as per \Cref{def:TM-RA}, robust accuracy measures stability rather than correctness since we aren't evaluating probability of the event $\left\{\forall\;  \bdel \in \mathcal{S}(\x, d, \varepsilon),~~h(\x+\bdel) = h^\star(\x+\bdel)\right\}$.} of a classifier's prediction under perturbations within the permissible sets. The robust supervised learning task seeks a classifier in $\cH$ with high robust accuracy. 

\begin{definition}[$\varepsilon$-robust] \label{def:eps-robust}
A classifier $h \in \cH$ is \emph{$\varepsilon$-robust} at $\x$ w.r.t. a threat function $d$ if, 
$
\forall ~ \bdel \in \mathcal{S}(\x, d, \varepsilon) \cap (\cX-\{\x\}), \text{ we have } h(\x + \bdel) = h^\star(\x).
$
\end{definition}
An $\varepsilon$-robust classifier predicts the same label $h^\star(\x)$ at $\x+\bdel$ for any corruption $\bdel$ within the $\varepsilon$-sublevel set $\mathcal{S}(\x, d, \varepsilon)$. We emphasize that robust accuracy and $\varepsilon$-robustness are measures of stability rather than correctness. 
In this way, each threat model $(d, \varepsilon)$ implicitly encodes a trade-off between stability and correctness \citep{Dohmatob2019GeneralizedNF}. Requiring more stability than necessary can lead to incorrect predictions as shown in \Cref{fig:imagenet-motivation-1}. 

\section{Threat specification beyond $\ell_p$ norms}

In this section, we design a principled threat model from first principles that is local and anisotropic. We first theoretically develop our exact PD$^\star$-threat model assuming knowledge of the partition sets $\cX_y$ of the true labeling function $h^\star$ (\Cref{subsec:exact}), and later relax this assumption by developing an approximation leveraging the observed training data and obtaining a practical PD threat (\Cref{subsec:observed}).

\subsection{Measuring threat with class partition}\label{subsec:exact}
\begin{figure}[t]
\centering
\includegraphics[width=0.5\linewidth]{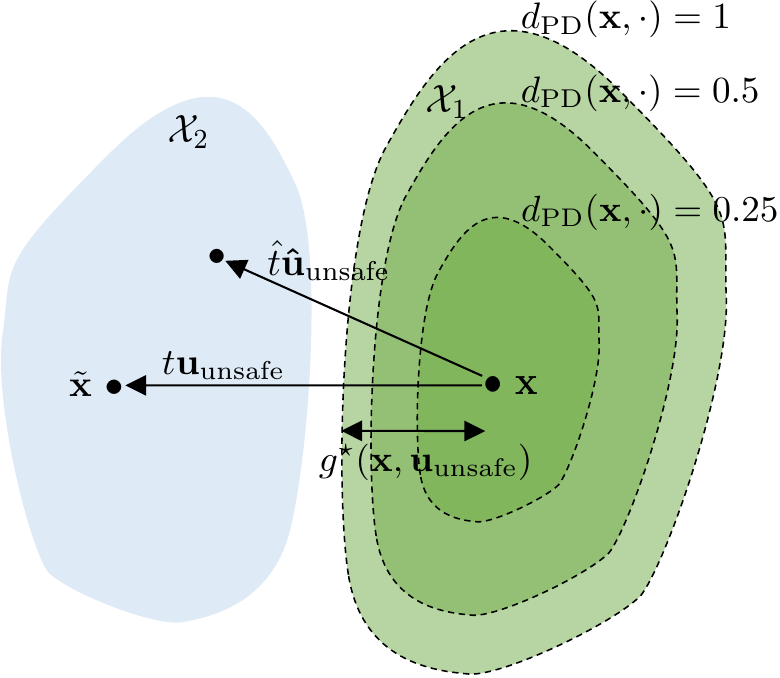}
\caption{An illustration of unsafe directions, and sub-level sets of the PD threat.}
\label{fig:levelsets}
\end{figure}
\begin{definition}[Unsafe Directions]\label{def:unsafe}
At each input $\x$, a direction $\vc{u} \in \mathbb{S}^d_2$ is called \textit{unsafe} if there exists a step size $t \geq 0$ such that $\x+t \vc{u} \in \cX$ and $h^\star(\x+t \vc{u}) \neq h^\star(\x)$. 
We denote the set of \textit{all} unsafe directions at $\x$ as $\mathcal{U}^\star(\x)$. 
\end{definition}
The minimum step size needed to alter the true label can vary across unsafe directions. 
Directions $\vc{v} \in \mathbb{B}^d_2 \cap (\mathcal{U}^\star(\x))^c$ are are called \textit{safe}, since moving along them (in the data domain) cannot alter the true label. We note that adversarial perturbations (\Cref{def:adv-rob}) are a characteristic of a learned predictor $h \in \cH$, while unsafe directions are a characteristic of the true labeling function $h^\star$. We propose to measure threat of a perturbations $\bdel$ using $\mathcal{U}^\star(\x)$.

\begin{definition}[PD$^\star$-threat]\label{def:noniso-threat}
Let $\x \in \cX$ and let $\bdel \in \mathbb{R}^d$ be a perturbation. The \textit{exact} projected displacement threat function $d^\star_{PD}$ is defined as the maximum scaled displacement of the Euclidean projection 
of the perturbation $\bdel$ over \textit{all} unsafe directions, 
\begin{alignat}{3}
    d^\star_{\mathrm{PD}}(\x, \bdel) \coloneqq&~~ \underset{\vc{u} \;\in\; \mathcal{U}^\star(\x)}{\sup} ~~ \frac{1}{g^\star(\x, \vc{u})} \max\left(\langle \bdel, \vc{u}\rangle, 0\right)  %
    \label{def:pdisp}
\end{alignat}
where $g^\star(\vc{u}, \x)$, called the \textit{normalization function}, is defined as 
\begin{align}
\label{eq:normfn}
g^\star(\x, \vc{u}) \coloneqq \;\;
\underset{\mathbb{R}^{\geq 0}}{\sup} ~ 
 & \quad M \\
\nonumber ~\text{s.t.}  ~  
&\quad 
\x + t\vc{u} \in \cX_{h^*(\x)}
 \;\; \forall\; t \in [0,M]. 
\end{align}

\end{definition}
\noindent
At each input $\x$, the normalization function $g^*(\x, \vc{u})$ captures the distance along the direction $\vc{u}$ to the boundary of the set $\cX_{h^\star(\x)}$. The normalization function enables a comparison of the threat along different unsafe directions (see \Cref{fig:levelsets}). In \eqref{def:pdisp}, the supremum ensures that we do not underestimate the threat along a direction that potentially alters the true label. We highlight two key characteristics of the exact projected displacement threat function $d^*_{\rm PD}$. 

\textbf{Anisotropy}.
The threat function $d^\star_{\rm PD}$ is anisotropic (the threat of a perturbation $\bdel$ at input $\x$ depends on both the direction and magnitude of the perturbation) and does not correspond to a norm in general, unlike $\ell_p$ threat models. For a fixed $\x$, the threat along different directions $\vc{u}$ varies based on the alignment with unsafe directions $\mathcal{U}^\star(\x)$. In particular, $d^\star_{\rm PD}$ can differentiate between the perturbations $\bdel$ and $-\bdel$. Anisotropy is critical for differentiating perturbations along safe and unsafe directions. 

\textbf{Locality}.
The threat function $d^\star_{\rm PD}$ exhibits locality; i.e., a fixed perturbation $\bdel$ has varying threat $d^\star_{\rm PD}(\x, \bdel)$ depending on the input $\x$. This means that the permissible set of perturbations $\mathcal{S}(\x, d^\star_{\rm PD}, \varepsilon)$ varies with the input\footnote{Unlike $\ell_p$ threat models where $\mathcal{S}(\x, d_p, \varepsilon) = \varepsilon\mathbb{B}^d_p \;\;\forall \; \x \in \cX$.}, enabling an input-aware assessment of threat. We note that weighted $\ell_p$ norms \cite{erdemir2021adversarialrobustnessnonuniformperturbations} can be anisotropic but not local.

\subsubsection{Illustration of PD*-threat $d^*_{\rm PD}$ on synthetic 2D data}
\begin{figure}[H]
\centering 
    \begin{minipage}{0.49\linewidth}
        \centering     
        \includegraphics[width=0.75\textwidth]{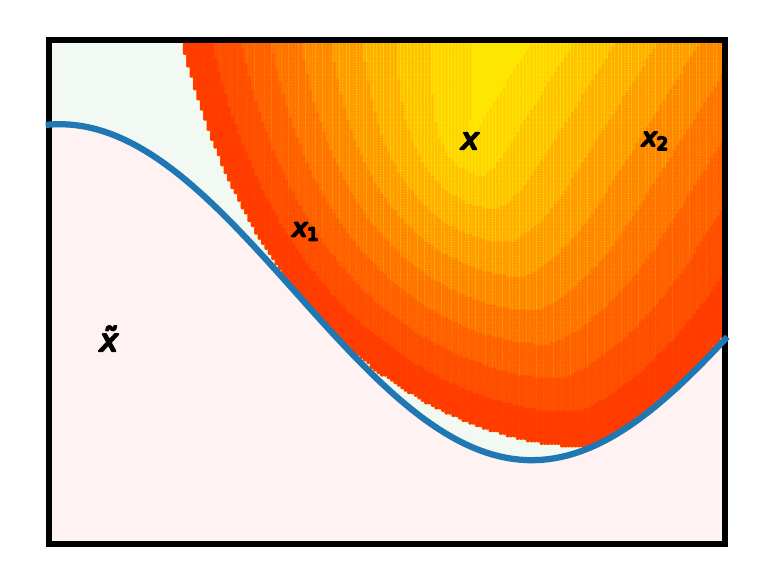}\\
        \subcaption{$\mathcal{S}(d^*_{\rm PD},\x, 1)$}
        \label{fig:synthetic-1}
    \end{minipage}\hfill
    \begin{minipage}{0.49\linewidth}
        \centering     
        \includegraphics[width=0.75\textwidth]{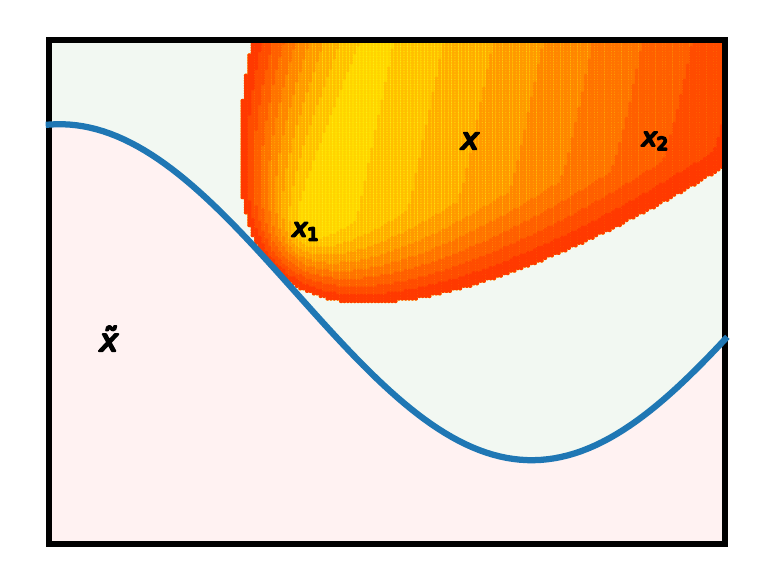}\\
        \subcaption{$\mathcal{S}(d^*_{\rm PD}, \x_1, 1)$}
        \label{fig:synthetic-2}
    \end{minipage}\\
    \begin{minipage}{0.49\linewidth}
        \centering     
        \includegraphics[width=0.75\textwidth]{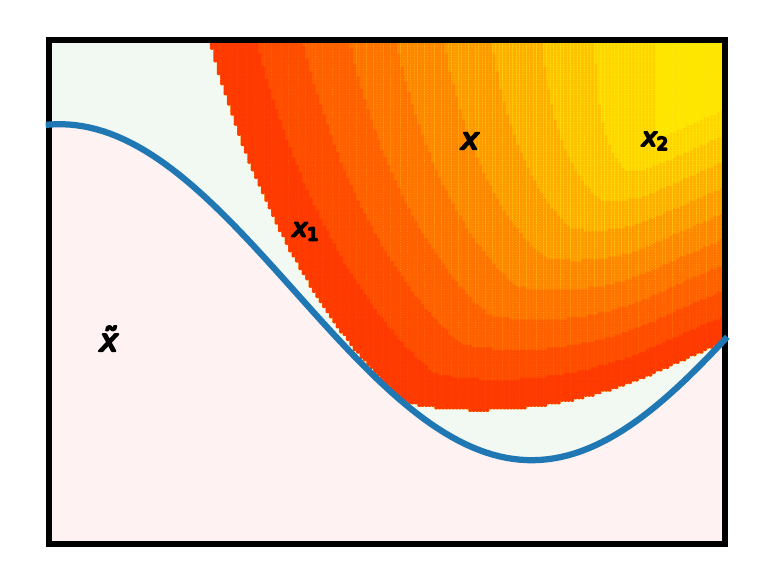}\\
        \subcaption{$\mathcal{S}(d^*_{\rm PD}, \x_2, 1)$}
        \label{fig:synthetic-3}
    \end{minipage}\hfill 
    \begin{minipage}{0.49\linewidth}
        \centering     
        \includegraphics[width=0.75\textwidth]{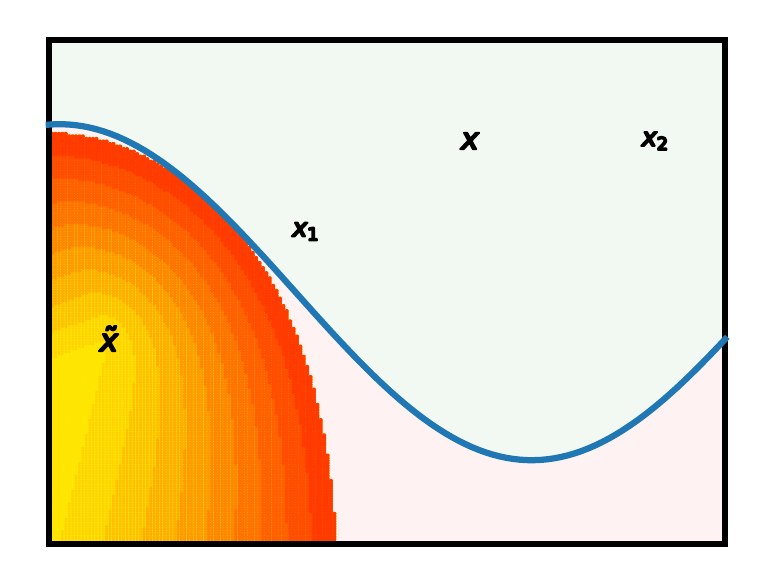}\\
        \subcaption{$\mathcal{S}(d^*_{\rm PD}, \xtil, 1)$}
        \label{fig:synthetic-4}
    \end{minipage}%
    \caption{Shape of 1-sublevel sets  at different inputs.}
\label{fig:synthetic}
\end{figure}

Consider a binary classification task where inputs $\x \in \mathbb{R}^2$ from a bounded domain $\cX$ are assigned labels $y \in \cY := \{-1, +1\}$ by a true labeling function $h^*$. In \Cref{fig:synthetic}, the solid black rectangular regions indicate the bounded domain $\cX$, and the solid blue lines are the decision boundary of the true labeling function $h^*$. A point within the domain is labeled $1$ if it is above the blue line and $-1$ otherwise. \Cref{fig:synthetic} presents four inputs, three points ($\x, \x_1, \x_2$) with label $1$ and $\xtil$ with label $-1$. The points are chosen such that $\x_1$ and $\x_2$ are equidistant from $\x$, i.e., $\norm{\x_1-\x}_2 = \norm{\x_2-\x}_2 = \varepsilon$. In this toy example, the true labeling function $h^*$ and the corresponding partition sets $\cX_{1}$ and $\cX_{-1}$ are known, and thus the exact PD threat $d^*_{\rm {PD}}$ can be computed\footnote{Unsafe directions $\mathcal{U}^*(\x)$ and normalization $g^*(\x, \vc{u})$ are computed via a 2D discretization grid over the domain $\cX$.}. At $\x$, clearly $\vc{u}_1:= \frac{\x_1-\x}{\norm{\x_1-\x}_2}$ is an unsafe direction while $\vc{u}_2 := \frac{\x_2-\x}{\norm{\x_2-\x}_2}$ is a safe direction and hence $d^*_{\rm PD}(\x, \varepsilon \vc{u}_1) \geq d^*_{\rm PD}(\x, \varepsilon \vc{u}_2)$.  \Cref{fig:synthetic} presents a visualization of the 1-sublevel sets at each marked point. 

\Cref{fig:synthetic} demonstrates the anisotropy and locality of the sublevel sets $\mathcal{S}(d^*_{\rm PD}, \cdot, 1)$.
A large value of threat $d^*_{\rm PD}(\x, \bdel)$ indicates the proximity of the input $\x$ to the boundary of the partition sets. This behaviour is intuitively captured in \Cref{fig:synthetic} where $\x_1$ is closer to the boundary indicated by the blue line than the other marked points.

\subsubsection{Stability vs Correctness Trade-off}
Since the normalization function $g^\star$ accounts for the class partition $\{\cX_{c}\}_{c=1}^C$, the threat function $d^\star_{\rm PD}$ has a principled trade-off between stability and correctness. Specifically, the true labeling function $h^\star$ exhibits a high level of robustness under the PD$^\star$-threat. 

\begin{theorem}\label{thm:1-rob}
The true labeling function $h^\star$ is $1$-robust at any input $\x\in \cX$ w.r.t. the threat $d^\star_{\mathrm{PD}}$. Additionally, if a classifier $h$ is not 1-robust at all inputs then there exists an input $\x$ misclassified by $h$.  
\end{theorem}
In other words, for any pairs of inputs $\x, \xtil \in \cX$ with different labels, the threat $d^\star_{\rm PD}(\x, \xtil-\x)$ is always larger than 1. Thus seeking $h\in \cH$ that is $1$-robust at any input under $d^\star_{\rm PD}$ is not at odds with learning $h^\star$. Proof of the above result can be found in \Cref{app:1-rob}.

\subsection{Measuring threats with observed data}\label{subsec:observed}
Computing $d^\star_{\rm PD}(\x, \cdot)$ from \cref{def:noniso-threat} requires explicitly characterizing the set of unsafe directions $\mathcal{U}^\star(\x)$ and the normalization function $g^\star(\vc{u}, \x)$, both of which need knowledge of the partition sets $\{\cX_{c}\}_{c=1}^C$ induced by $h^\star$. We now discuss an empirical approximation based on observed training data.

\textbf{Approximating $\mathcal{U}^\star(\x)$}. We have access to training data $\samp$ and incomplete information on the partition sets $\{\cX_{c}\}_{c=1}^C$ via the partition $\samp = \cup_{c=1}^C \samp_c$, where $\samp_c \coloneqq \{\x \in \cX ~|~ (\x, c) \in \samp\} \subset \cX_{c}$ contains observed training data with class label $c$. At every training input $\x \in \samp_y$ and for each input $\xtil \in \samp_c$ (for $c\neq y)$, the direction $\vc{u} = \frac{\xtil-\x}{\norm{\xtil-\x}_2}$ is an unsafe direction. For computational considerations, we choose a collection $\bigcup_{c} \samp_{c,k}$ of representative subsets $\samp_{c,k} \subset \samp_c$ of size $k$. As a heuristic choice, we select subsets $\samp_{c,k}$ by solving a discrete-k-center optimization problem (detailed in \cref{sec:ksubset}). We are now ready to define an observed subset of the unsafe directions based on $\samp_{c,k}$.

\begin{definition}[Observed Unsafe Directions]
\label{def:unsafe-observed-k}
For each $\x \in \samp_y$, we identify a set of $(C-1)k$-observed unsafe directions based on the representative subsets $\samp_{c,k}$,
\begin{align*}
\mathcal{U}_k(\x) \coloneqq 
\left\{ 
\frac{\xtil-\x}{\norm{\xtil-\x}_2}
~\Big|~ 
\xtil \in \underset{c \neq y}{\cup} \samp_{c,k}
\right\} 
\subset \mathcal{U}^\star(\x) .
\end{align*}  
\end{definition}

\paragraph{Approximating $g^\star(\x, \vc{u})$.} 
At any input $\x \in \cX_y$, for each observed unsafe direction $\vc{u} \in \mathcal{U}_k(\x)$, there exists an observed training point $\xtil \in \cup_{c \neq y} \samp_{c,k}$ such that $\vc{u} = \frac{\xtil-\x}{\norm{\xtil-\x}_2} \in \mathcal{U}_k(\x)$. The normalization function \eqref{eq:normfn} at input $\x$ and direction $\vc{u}$ can be bounded as $g^\star(\x,\vc{u}) \leq \|\xtil - \x\|_2$ since $h^\star(\xtil) \neq h^\star(\x)$. In practice, we use the heuristic $g_{\beta}(\x, \vc{u}) \coloneqq \beta\norm{\xtil-\x}_2$, where $\beta \in (0,1)$ is a scaling hyper-parameter to compute an approximate normalization along the observed unsafe directions $\vc{u}\in \mathcal{U}_k(\x)$.

\begin{definition}[$(k,\beta)$-PD threat]\label{def:noniso-threat-k}
Let $\x \in \cX$ and let $\bdel \in \mathbb{R}^d$ be a perturbation. The projected displacement threat is defined as
\begin{alignat}{3}
    d_{\mathrm{PD}, k, \beta}(\x, \bdel) \coloneqq&~~ \underset{\vc{u} \;\in\; \mathcal{U}_k(\x)}{\max} ~~ \frac{1}{g_{\beta}(\x, \vc{u})} \max\left(\langle \bdel, \vc{u}\rangle, 0\right). \label{def:pdisp-k}
\end{alignat}
\end{definition}
We first note that the approximation $d_{\rm PD, k, \beta}$ of the exact threat $d^\star_{\rm PD}$ inherits the anisotropy and locality property. We now state a few additional important properties of $d_{\rm PD}$. 

\textbf{Quality of Approximation}. 
The quality of the approximation depends on the choice of representative unsafe directions $\mathcal{U}_{k}(\cdot)$ via the $k$-subset $\samp_{c,k}$ and on the heuristic choice of the approximate normalization $g_{\beta}$, via the scaling hyper-parameter $\beta$. 
For approximating the normalization, we note that for $\beta=1$, $d_{\rm PD, k, 1}(\x, \bdel) \leq d^*_{\rm PD, k, \beta}(\x, \bdel)$ for all inputs $\x$ and perturbations $\bdel$. However, the permissible set $\mathcal{S}(\x, d_{\rm PD, k,\beta}, \varepsilon)$ for threshold $\varepsilon=1$ is likely to include unsafe perturbations. To see this, let $\vc{u} \in \mathcal{U}_k(\x)$ and let $\xtil \in \samp_{c,k}$ be the corresponding point such that $\vc{u}:=\frac{\xtil-\x}{\norm{\xtil-\x}_2}$. The perturbation $\bdel := \xtil-\x$ has threat $d_{\rm PD, k, 1}(\x, \bdel)=1$. Yet, for some $t \leq 1$, the scaled perturbation $t\bdel$ has threat less than 1 but $\x+t\bdel$ still has true label $c$ and hence $t\bdel$ is an unsafe perturbation. Hence, in practice, we make a heuristic choice of $\beta=\frac{1}{2}$ to compute the approximate normalization. 

Next 
we note that a larger $k$ trades-off computational efficiency of evaluating $d_{\rm PD, k,\beta}$ to how well it approximates $d^*_{\rm PD}$. Given training data $\samp$, we first recommend finding the minimum $k$ such that PD threat $d_{\rm PD, k,\beta}$ rates unsafe corruptions on training data as a sufficiently large threat, 
\[
k_{\rm min} :=  \min_{k \in [1, \frac{|S|}{C}]} 
\text{ s.t. } \min_{(\x,y),\; (\xtil,c) \in \samp} d_{\rm PD, k, \beta} (\x, \xtil-\x)  > 1
\]
We let $k_{\rm max} \in [k_{\rm min}, \frac{|S|}{C}]$ be the maximum $k$ subject to a practitioner's memory constraints and desired throughput on evaluation of threat function. We then recommend a equi-spaced grid search over the interval $[k_{\rm min}, k_{\rm max}]$ to determine an appropriate heuristic choice. In our experiments on Imagenet-1k, we observed that $k_{\rm min} = 20$ and $k_{\rm max}=50$ with our computing resources. In this range, a grid search showed that $k = 50$ is sufficiently fast, and capable of discriminating safe and unsafe perturbations on validation data (more on this in \Cref{Sec:Experiments}). In this article we fix $(k,\beta)=(50,\frac{1}{2})$ as default hyper-parameters and, for brevity, refer to the observed unsafe directions (\Cref{def:unsafe-observed-k}), the heuristic normalizations and the PD threat (\Cref{def:noniso-threat-k}) as $\mathcal{U}(\x), g(\x,\bdel)$ and $d_{\rm PD}(\x,\bdel)$, respectively. 

\textbf{Growth and Sensitivity.}
The growth of PD threat along any direction is linear, i.e., $d_{\rm PD}(\x, t \bdel) = t d_{\rm PD}(\x, \bdel)$, and the rate of growth scales inversely with the (approximate) normalization. In particular, for a fixed  $\x$, $d_{\rm PD}(\x, \cdot)$ is $\left(\max_{u \in \mathcal{U}(\x)}1 / g(\x,\vc{u})\right)$-Lipschitz w.r.t perturbations in the $\ell_2$ norm. 

\textbf{Projection onto sublevel sets.}
For any threat model $(d, \varepsilon)$, the ability to efficiently project a perturbation $\bdel$ onto the sub-level sets $\mathcal{S}(d, \x, \varepsilon)$ enables one to leverage a wide literature on gradient-based attacks to conduct a rigorous evaluations of adversarial robustness. The sublevel sets of PD threat model, i.e., $\mathcal{S}(\x, d, \varepsilon)$, are convex for all $\x$ for all $\epsilon \geq 0$. In particular, $\mathcal{S}$ is the intersection of at most $k \cdot C$ half-spaces characterized by unsafe directions $\mathcal{U}(\x)$, as $\mathcal{S}(\x, d_{\rm PD}, \varepsilon) = 
\underset{\vc{u} \in \mathcal{U}(\x)}{\cap} \{\bdel \in \mathbb{R}^d ~|~  \langle \bdel, \vc{u} \rangle \leq \varepsilon\cdot g(\x,\vc{u}) \}
$. Further, the sub-level sets are monotonic w.r.t. the threshold, i.e., for $\varepsilon_1 \geq \varepsilon_2 \geq 0$, $\mathcal{S}(\x, d_{\rm PD},\varepsilon_2) \subseteq \mathcal{S}(\x, d_{\rm PD}, \varepsilon_1)$. 

Linearity of growth and convexity of the sub-level sets together provide a straightforward approximate projection algorithm, since for any perturbation $\bdel$, the scaled perturbation $\frac{\varepsilon}{d_{\rm PD}(\x, \bdel)} \bdel$ lies in the permissible set $\mathcal{S}(\x, d_{\rm PD}, \varepsilon)$. Further, due to the convexity of sublevel sets, an exact projection (in $\ell_2$ distance)  can be computed using the iterative greedy scheme outlined in the \Cref{alg: greedyproject} (deferred to the appendix). Thus the PD threat provides the practitioner with the flexibility of a fast approximate projection or an exact projection at the expense of increased computation. Together, these projection algorithms enable immediate adaptation of existing adversarial attacks to our proposed threat model. 

We note that efficient exact projection algorithms are also known for $\ell_p$ threat models. 
However, for neural perceptual threat models such as DreamSim \citep{fu2024dreamsim}, the sub-level sets are non-convex and a projection can only be approximately computed \citep{laidlaw2021perceptualadversarialrobustnessdefense} and is typically not computationally efficient.

\section{Experiments on Real World Data}\label{Sec:Experiments}
In this section we illustrate the qualitative and quantitative characteristics of the PD threat in comparison to existing threat models. We evaluate 4 threat functions: $d_{\infty}, d_2, d_{\rm DS}$, and our proposed PD threat function $d_{\rm PD}$, where $d_{\rm DS}$ is the neural perceptual threat function $d_{\rm DS}(\x, \xtil-\x) \coloneqq \textrm{DreamSim}(\x, \xtil)$ from DreamSim \citep{fu2024dreamsim}. Note that we focus on DreamSim due to its superior performance over other neural perceptual threat models \cite{fu2024dreamsim}, but the following comparison applies to any modern neural perceptual threat model. %

\textbf{Experimental Setup}. We seek to measure the ability of threat functions to distinguish between safe and unsafe corruptions. The threat models $d_{\rm PD}$ and $d_{\rm DS}$ depend on the Imagenet-1k training dataset $\samp$. All threat models are evaluated on the Imagenet-1k validation dataset $\samp_{\mathrm{val}}$. For any pair of labelled inputs $(\x,y) , (\xtil,c) \in \samp_{\rm val}$ with distinct labels ($c\neq y$), the perturbation $\bdel:=\xtil-\x$ is an \emph{unsafe} perturbation at $\x$. This way, we can evaluate the statistics of threat under unsafe perturbations for each threat function. To characterize threat under safe perturbations, we consider a set $\Omega$ of 150 distinct common corruptions consisting of 30 styles $\times$ 5 severity levels, i.e., severity $\in \{1,2,3,4,5\}$ -- the supplementary material contains visual examples (see \Cref{fig:lionfish_corruptions_c,fig:lionfish_corruptions_c_bar}), sourced from \textbf{Imagenet-$C$} \citep{hendrycks2019robustness} and \textbf{Imagenet-$\bar{\text{C}}$} \cite{NEURIPS2021_1d497805} accompanied with their threat assessment. 
\begin{table}[t!]
    \centering
    \resizebox{\columnwidth}{!}{%
    \begin{tabular}{|c|c|} 
    \toprule
       \textsc{Category}  & \textsc{Corruption Styles}\\
    \midrule
       Noise  & {\small Gaussian, Shot, Impulse, Speckle, Blue, Brown, Perlin, Single-Frequency, Plasma}\\  
       \midrule
       Blur & {\small Gaussian, Defocus, Glass, Motion, Concentric Sine Waves, Caustic Refraction}\\
       \midrule
       Compression & {\small Pixelate, JPEG}\\
       \midrule
       Digital & {\small Brightness, Contrast, Saturate, Elastic Transform}\\
       \midrule
       Weather & {\small Frost, Fog, Snow}\\
       \midrule
       Occlusion & {\small Spatter, Checkerboard Cutout, Sparkles, Inverse Sparkles}\\
    \bottomrule
    \end{tabular}
    }
    \caption{Categories of corruptions studied, from \textbf{Imagenet-C} \citep{hendrycks2019robustness} and \textbf{Imagenet-$\bar{\text{C}}$} \cite{NEURIPS2021_1d497805}.}
    \label{tab:corruption_groups}
\end{table}

For a specific corruption style such as Gaussian noise and a severity level $5$, we let $\Omega_{\rm Gaussian Noise, 5}$ denote the corresponding corruption function. To aid visualization, we group the corruption styles into 6 distinct categories (listred in \Cref{tab:corruption_groups}),  
\[
\displaystyle \Omega := \cup_{i\in [5]}~ \Omega_{i} := \cup_{i\in [5]}~ \Big(\;\Omega_{\rm noise, i} \cup \Omega_{\rm blur, i} \cup \Omega_{\rm compression, i} \cup \Omega_{\rm digital, i} \cup \Omega_{\rm weather, i} \cup \Omega_{\rm occlusion, i}\;\Big).
\] 
where $\Omega_i$ is the set of all corruption styles at severity level $i$. Each individual corruption $\omega \in \Omega$ is applied to a subset $\bar \samp \subset \samp_{\rm val}$ of 5,000 images chosen uniformly at random. For each threat function $d \in \{d_{\infty}, d_2, d_{\rm PD}, d_{\rm DS}\}$ and corruption $\omega \in \Omega$, for each image $\x \in \bar \samp$, we denote by $\mathrm{avg}(d, \omega)$, the average threat statistic, 
\[
\mathrm{avg}(d, \omega) := \frac{1}{|\bar{\samp}|} \sum_{(\x,y) \in \samp} d(\x, \omega(\x)-\x), 
\]
where $\omega(\x)$ is the corrupted input and $\omega(\x) - \x$ is the safe perturbation at $\x$. 

\subsection{Anisotropy of PD threat}
\begin{figure}
    \centering
    \begin{minipage}{0.49\linewidth}
        \centering     
        \includegraphics[width=\textwidth,trim={1.0cm 2.5cm 0.7cm 2.5cm},clip]{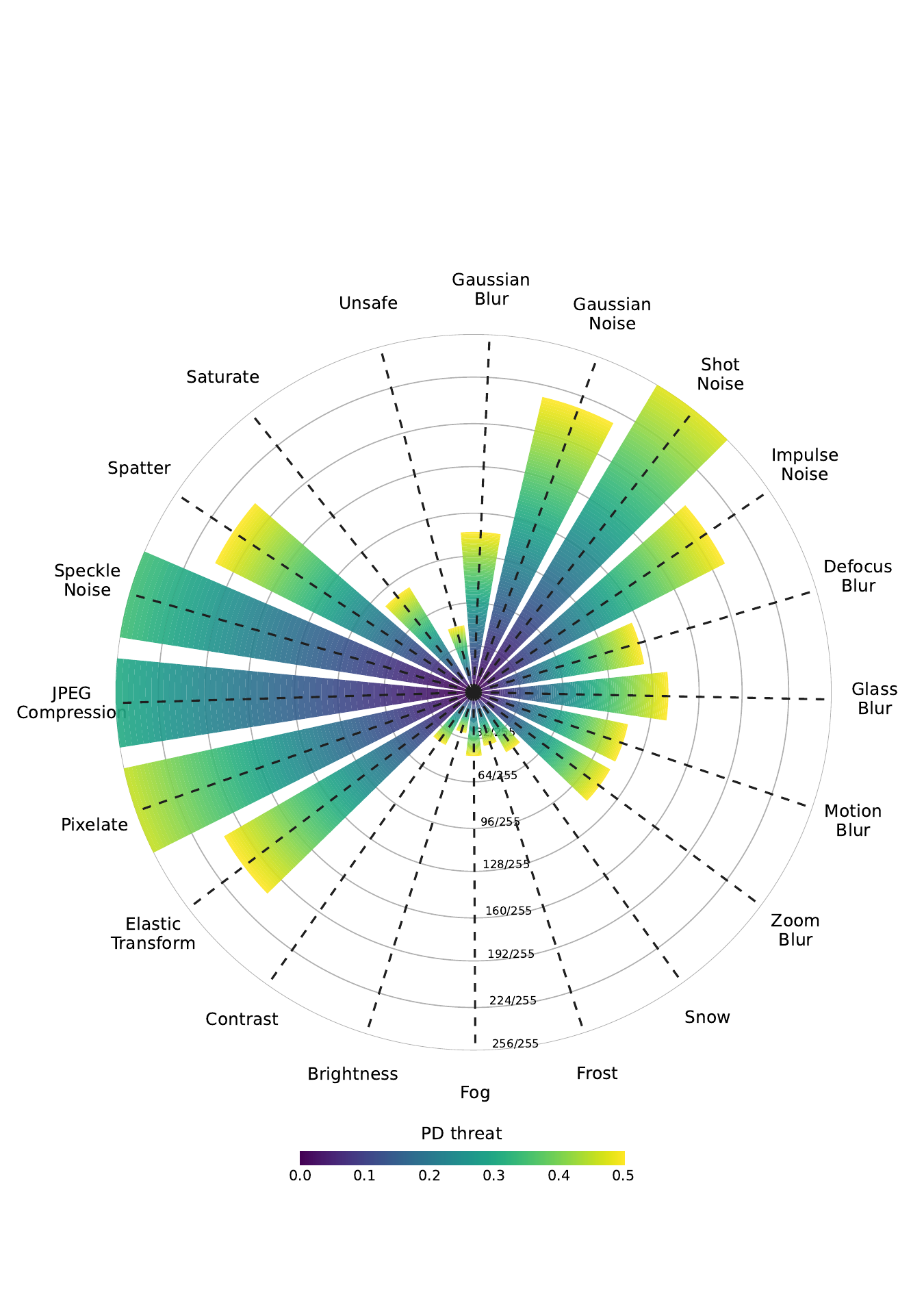}
        \subcaption{Imagenet-$C$}
        \label{fig:imagenet-5-1}
    \end{minipage}\hfill%
    \begin{minipage}{0.49\linewidth}
        \centering     
        \includegraphics[width=\textwidth, trim={1.0cm 2.5cm 0.7cm 2.5cm},clip]{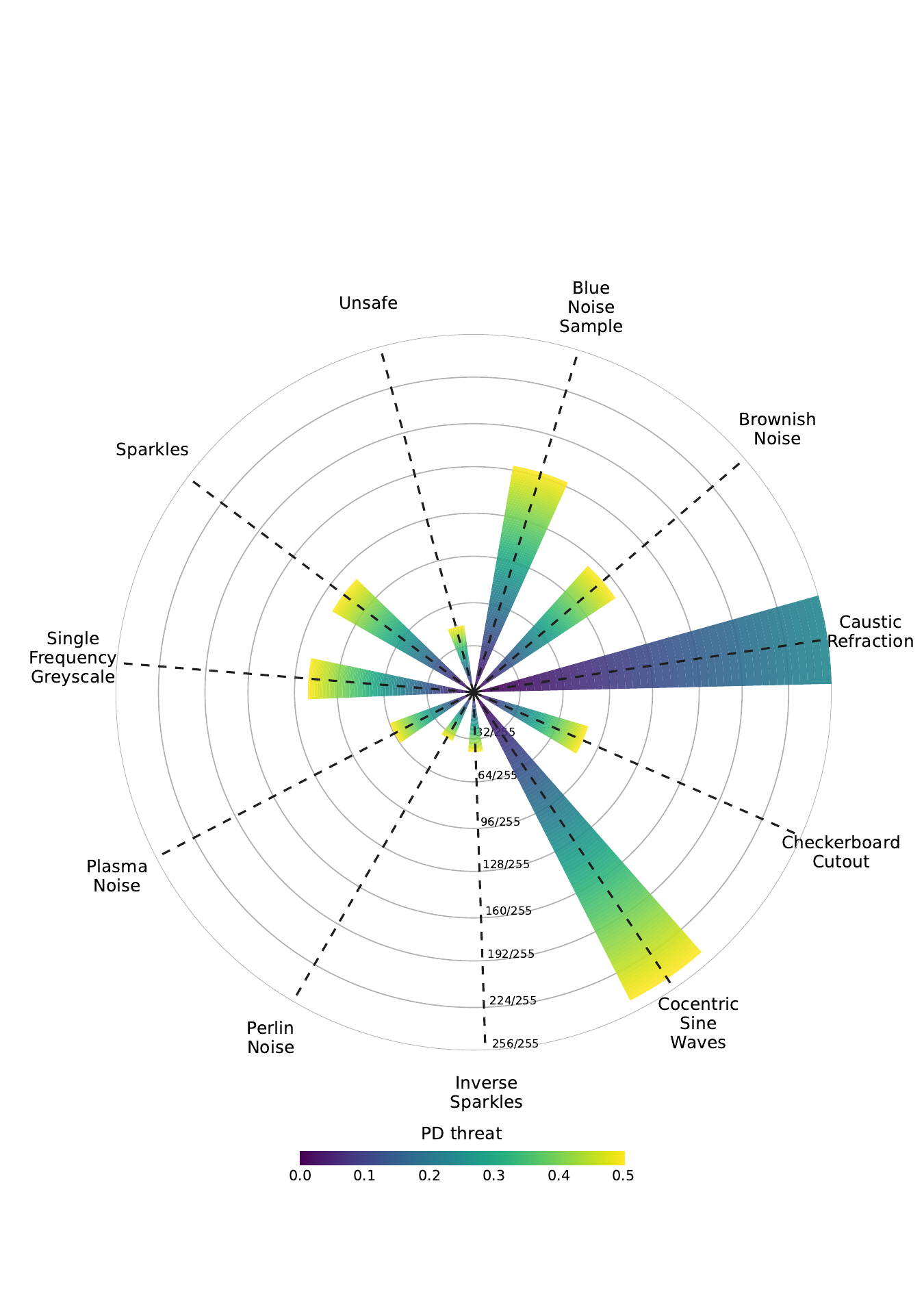}
        \subcaption{Imagenet-$\bar{C}$}
        \label{fig:imagenet-5-2}
    \end{minipage}\hfill 
    \caption{Anisotropy of PD threat Model}
    \label{fig:anisotropy-PD}
\end{figure}

At each input $\x$, along any direction $\vc{u}$, we can compare two threat functions $d_1$ and $d_2$ by measuring the largest $d_{1}$ threat for perturbations within $\mathcal{S}(d_{2}, \x, \varepsilon)$. Such a measurement is feasible for PD threat and $\ell_p$ threat due to linear growth, since, for any perturbation, evaluating the threat at $d(\x, \bdel)$, immediately provides the threat at scaled perturbations since $d(\x, t\bdel) = td(\x, \bdel)$ for each $d \in \{d_{\infty}, d_2, d_{\rm PD}\}$. \Cref{fig:anisotropy-PD} is a radial bar plot of the corruptions $\omega \in \Omega_5$ (with severity level 5) where the radial axis is $d_{\infty}$ threat. The heights of each radial bar is $\frac{1}{2 \cdot \mathrm{avg}(d_{\rm PD}, \omega)}\cdot \mathrm{avg}(d_{\infty}, \omega)$. A larger height indicates a corruption $\omega$ where the growth of PD threat per unit $\ell_{\infty}$ threat is lower (on average). \Cref{fig:anisotropy-PD} indicates that if corruptions $\omega(\x)-\x$ are scaled to a fixed $d_{\infty}$ threat, the resulting PD threat varies across directions reflecting the anisotropy of the PD threat model. We emphasize such a plot is not possible for the DreamSim threat model since the growth is non-linear and hence threat at each scaled perturbation $t\bdel$ needs to be evaluated separately. 

\subsection{Disentangling Safe and Unsafe Corruption}
\begin{figure}
    \centering
    \begin{minipage}{0.45\linewidth}
        \centering     
            \hspace{1.25cm} \includegraphics[width=0.7\linewidth]{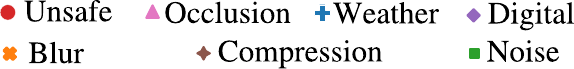} \\
        \includegraphics[width=\textwidth]{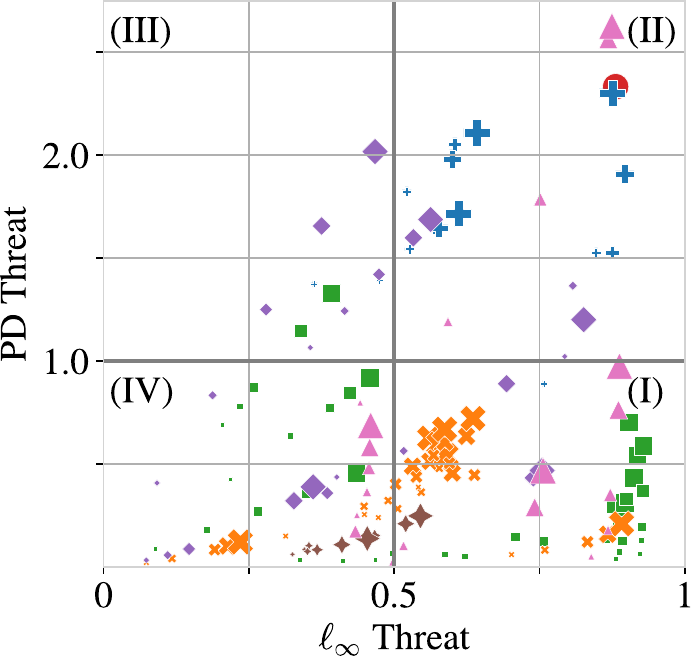}
        \subcaption{}
        \label{fig:PD-vs-Linf}
    \end{minipage}%
    \hspace{1cm}
    \begin{minipage}{0.45\linewidth}
        \centering     
            \hspace{1.25cm} \includegraphics[width=0.7\linewidth]{Figures/legend_corruptions.pdf} \\
        \includegraphics[width=\textwidth]{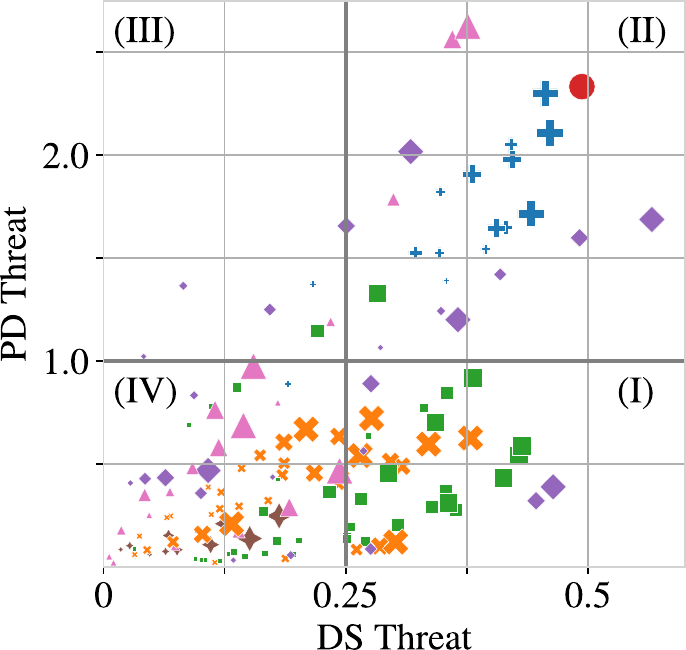}
        \subcaption{}
        \label{fig:PD-vs-DreamSim}
    \end{minipage}%
    \caption{Comparison of Threat models w.r.t. various common corruptions. The size of a marker represents the \emph{severity} level of the corruption. %
    }
    \label{fig:threat-comparison}
\end{figure}

We now measure the ability of threat models to distinguish safe and unsafe corruptions. Each corruption $\omega \in \Omega$ defines a point in \Cref{fig:PD-vs-Linf} having as the horizontal coordinate $\mathrm{avg}(d_{\infty}, \omega)$ and $\mathrm{avg}(d_{\rm PD}, \omega)$ as the vertical coordinate, producing a comparison of our PD threat to the $\ell_\infty$-threat model. Similarly, \Cref{fig:PD-vs-DreamSim} compares PD threat to the DreamSIM-threat model.

A specification of threat model $(d, \varepsilon)$ corresponds to the permissible set $\mathrm{S}(\cdot, d, \varepsilon)$ at each input. Such a choice is represented by a horizontal line for $d_{\rm PD}$ or a vertical line for ($d_{\infty}$ or $d_{\rm DS}$) in \Cref{fig:threat-comparison}. We note that a size-able set of common corruptions incur large $d_\infty$ threat, in particular as large as the unsafe corruptions (as indicated in \Cref{fig:imagenet-motivation-1}), thus any vertical line in \Cref{fig:PD-vs-Linf} that excludes the unsafe perturbations necessarily excludes several safe perturbations like the common corruptions. In contrast, both $d_{\rm PD}$ and $d_{\rm DS}$ admit choices of $\varepsilon$ that exclude unsafe perturbations while including most common corruptions. Hence, PD threat and DS threat are able to disentangle safe and unsafe corruptions. 

For each of the comparisons in \Cref{fig:threat-comparison}, we identify 4 quadrants (I-IV), distinguishing between low-threat and high-threat corruptions as measured by the corresponding threat models. We define the quadrants by choosing thresholds of $1, 0.5$, and $0.25$ for $d_{\rm PD}$, $d_\infty$, and $d_{\rm DS}$, respectively\footnote{These thresholds were chosen to approximately be half of the average threat of unsafe perturbations that alter the true label. Admittedly, this binary threshold is coarse, and the Appendix contains fine-grained qualitative examples of corruptions of varying threat.}, and comment on the qualitative behavior observed:
\begin{itemize}
    \item Quadrant IV contains corruptions that are characterized as low-threat by both threat models. As expected, almost all corruptions having a low severity ($1$) lie in this region. However, we already see deficiencies of the $d_\infty$ threat, as some blur, noise, and occlusion corruptions of severity $1$ in \Cref{fig:PD-vs-Linf} are not characterized as low-threat by at least one of the threat models. %
    \item Quadrant II contains corruptions that are characterized as high-threat by both threat models. These include digital and weather corruptions of high severity ($5$) as they tend to remove most of the signal in the input. 
    \item Quadrant I contains corruptions having low PD threat but high $d_\infty$ or $d_{\rm DS}$ and showcases the resilience of our PD threat model to natural corruptions like blur and Gaussian noise, which retain a lot of the information in the image relevant to the classification task. Indeed, a visual inspection of these corruptions (see  \Cref{appendix:illustrative} in the Appendix) reveals that even at a high severity, a human is able to discern task-relevant information from the corrupted images, and hence these corruptions should be assigned a low threat.
    \item Quadrant III contains corruptions with high PD threat but low $d_\infty$ or $d_{\rm DS}$ threat. We note that corruptions rated as high threat by $d_{\rm PD}$ are largely also rated highly by $d_{\rm DS}$. Since the PD and DS threats are aligned, the corruptions in Quadrant III in \Cref{fig:PD-vs-Linf} indicate that low $d_\infty$ is not sufficient to characterize hard corruptions.  
\end{itemize}

\begin{figure}[h]
	\centering
    \begin{minipage}{0.45\linewidth}
    \flushleft
        \includegraphics[width=\textwidth,trim={2.2cm 1cm 3cm 0.2cm},clip]{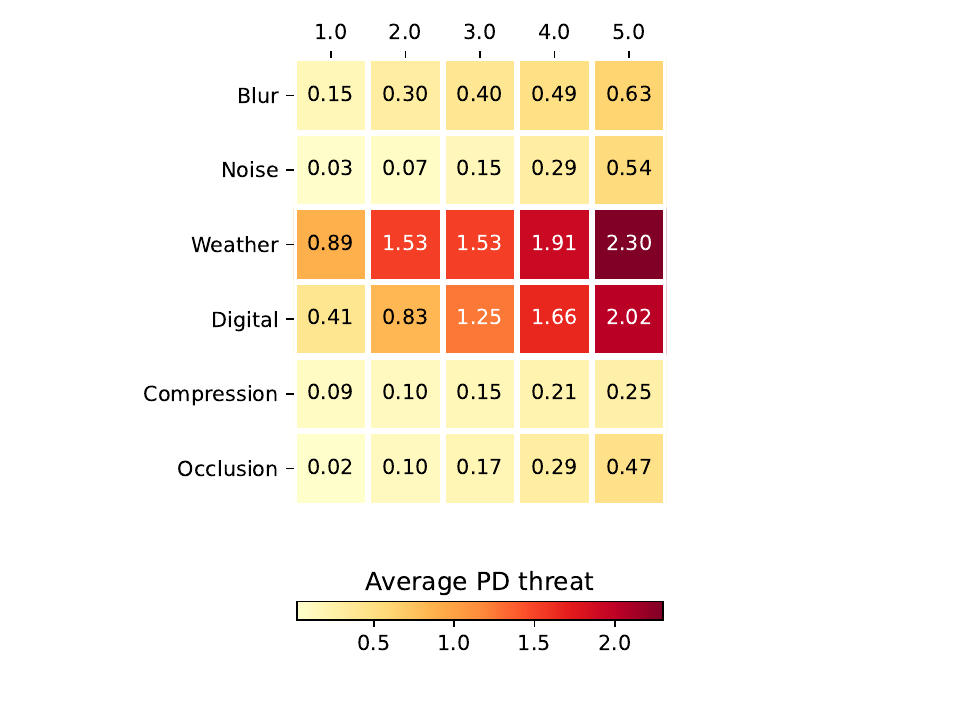}
        \label{fig:severity-PD}
    \end{minipage}%
    \begin{minipage}{0.45\linewidth}
    \centering
        \includegraphics[width=\textwidth,trim={2.2cm 1cm 3cm 0.2cm},clip]{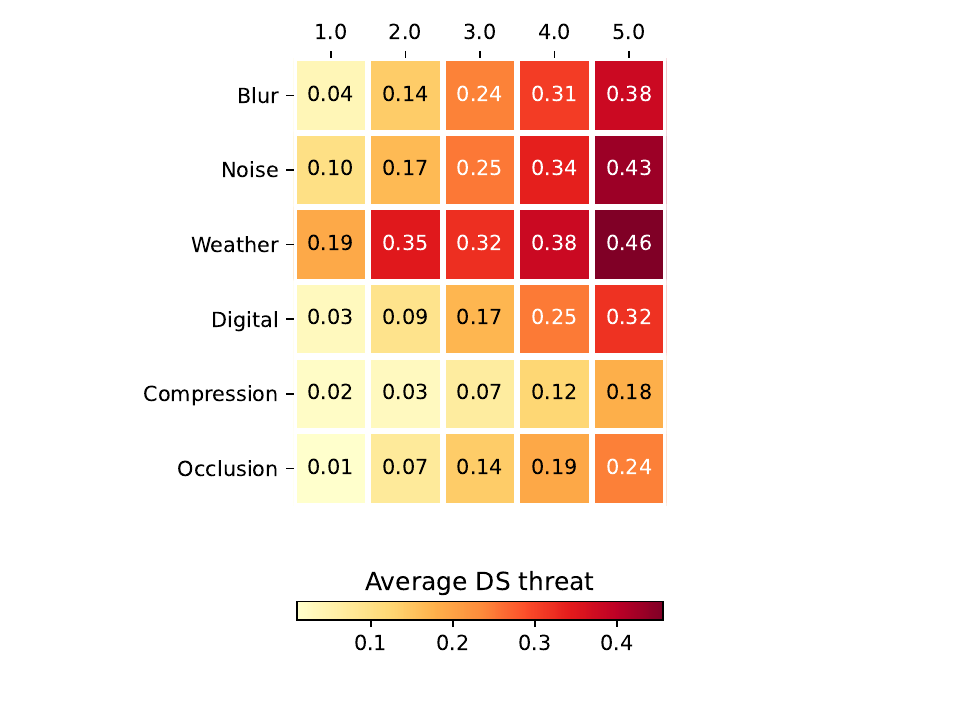}
        \label{fig:severity-DS}
    \end{minipage}%
    \caption{Heatmap of threat models vs severity levels of corruption groups}
    \label{fig:severity-comparison-PD}
\end{figure}

While the size of markers in \Cref{fig:threat-comparison} indicate the severity levels, different corruption styles are qualitatively different. For e.g, weather corruptions of severity level $i$ are uniformly harder than noise corruptions such as shot noise or Gaussian noise of a similar severity level. \Cref{fig:severity-comparison-PD} illustrates heatmaps of average threat for each corruption category across increasing severity. 

We observe that both PD and DS threat exhibit largely monotonic increase across severity levels, unlike $d_{\infty}$ threat. At this point, we will pause to summarize the benefits of PD threat. %

\textbf{PD vs DreamSim}. The PD threat compares well to the DS-threat but 
does not require pre-trained large vision models or fine-tuning on a curated dataset of human perceptual similarity judgements. For real-world classification tasks where perceptual similarity between pairs of input is harder to quantify even with expert human annotators,  instantiation of high-quality neural perceptual distance metrics is harder. %
Further, unlike perceptual distance metrics \cite{8578166, laidlaw2021perceptualadversarialrobustnessdefense, ghazanfari2023rlpips, Ghazanfari2023LipSimAP, fu2024dreamsim}, the growth of the PD threat is linear and interpretable along any direction, the sub-level sets induced by the PD threat model are convex, and admit efficient projection algorithms and hence PD threat presents a promising alternative. 

\section{Measuring threat with Task Annotation}\label{Sec:TaskAnnotation}

We described two practical design choices for the PD threat in \Cref{subsec:observed}: (1) a choice of the set of unsafe directions $\mathcal{U}(\x)$ at each input, and (2) a choice of normalization $g(\x, \vc{u})$ for each unsafe direction. Next, we incorporate additional task-relevant information to refine PD threat further.%

\subsection{Segmentation-aware Threat Specification}
Segmentation masks isolate regions of the input that contain semantic information relevant to the class label, and can be generated efficiently following recent advances in foundational models for image segmentation \cite{kirillov2023segany, ravi2024sam2}. We demonstrate a straightforward method to incorporate semantic masks in the assessment of threat. %

\begin{figure}[H]
\centering 
    \begin{minipage}{0.3\linewidth}
        \centering     
        \includegraphics[width=\textwidth]{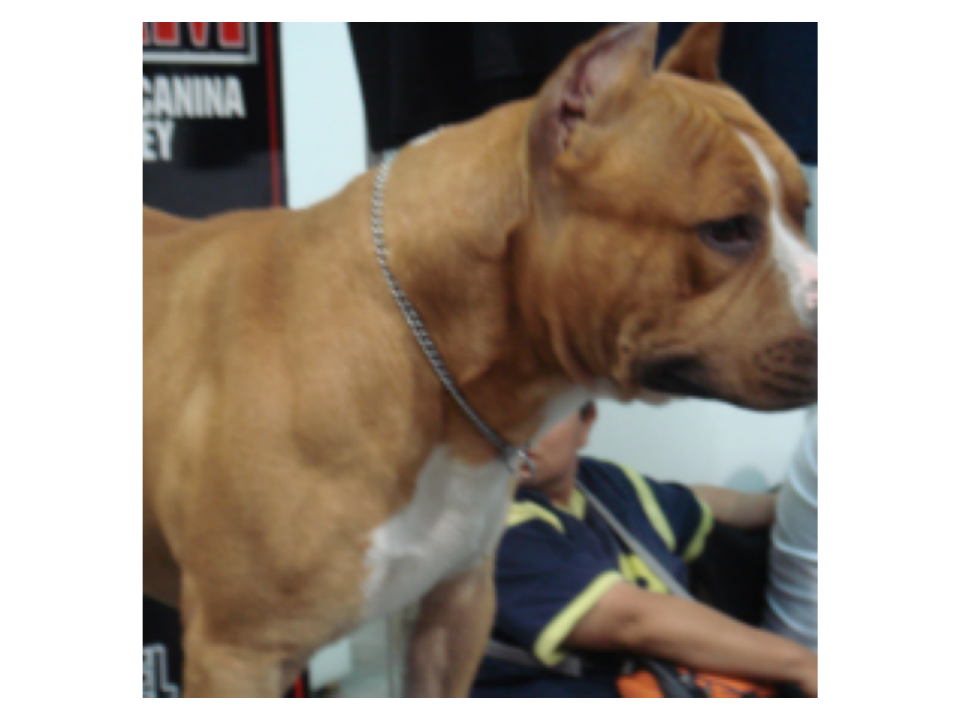}\\
        \subcaption*{\tiny Original Image \\ ~}
        \label{fig:PDS-1}
    \end{minipage}\hfill 
    \begin{minipage}{0.3\linewidth}
        \centering     
        \includegraphics[width=\textwidth]{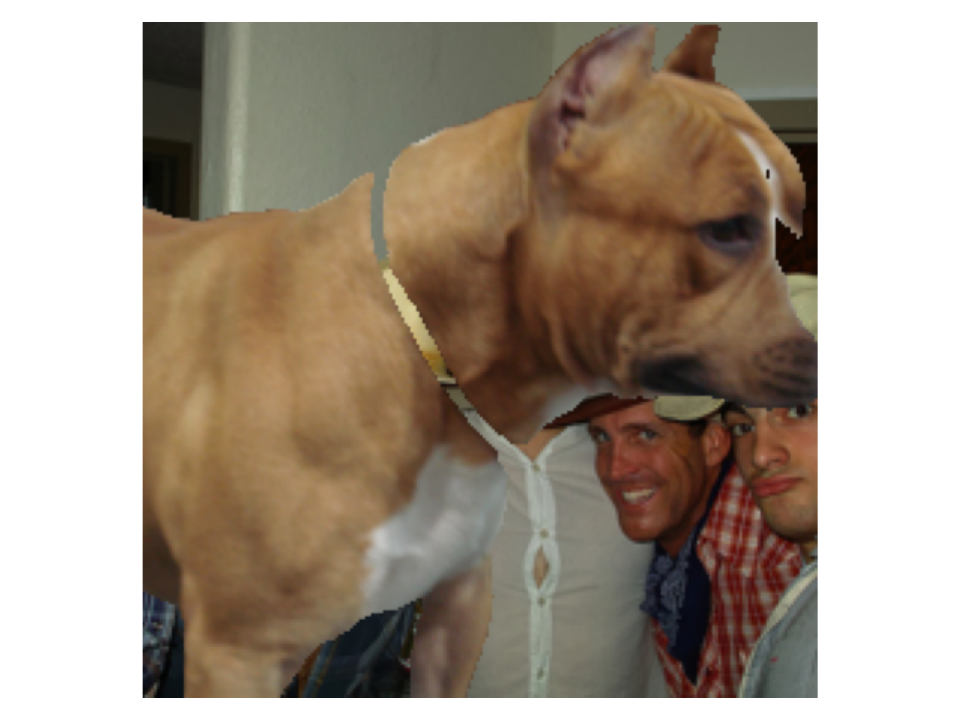}\\
        \subcaption*{\tiny PD = 1.4 \\ PD-S = 0.0}
        \label{fig:PDS-2}
    \end{minipage}\hfill
    \begin{minipage}{0.3\linewidth}
        \centering     
        \includegraphics[width=\textwidth]{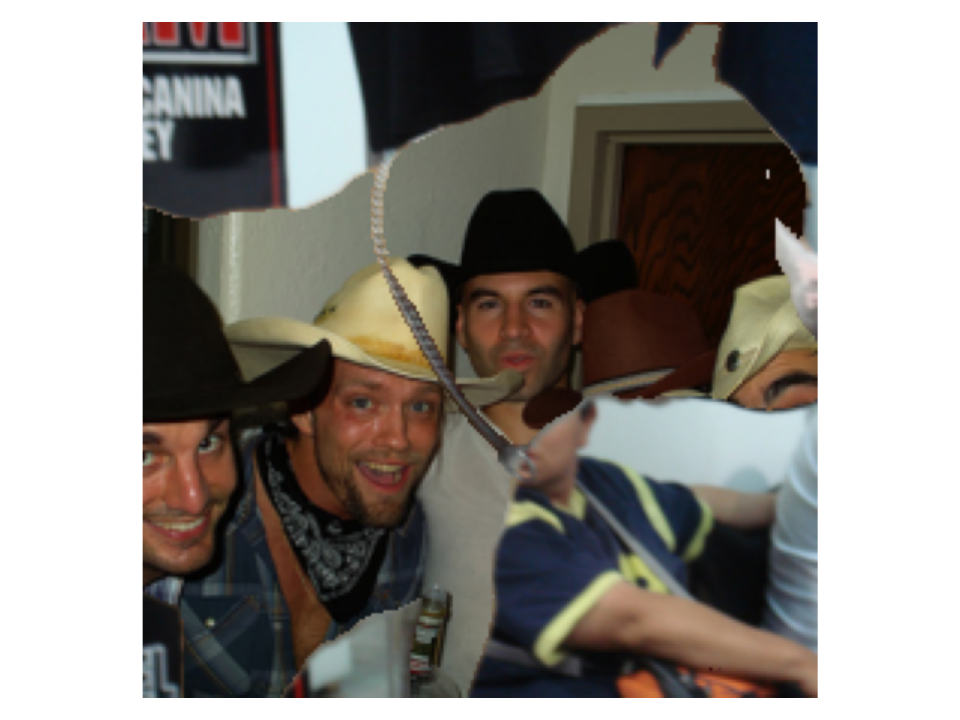}\\
        \subcaption*{\tiny PD = 0.68 \\ PD-S = 1.78}
        \label{fig:PDS-3}
    \end{minipage}\hfill
    \caption{Standard Projected Displacement (PD) threat vs Segmented Projected Displacement (PD-S) threat on background vs foreground corruptions}
\label{fig:PD-S}
\end{figure}

\begin{definition}[PD-S threat]\label{def:noniso-threat-k-segment}
Let $\x \in \cX$ and let $\bdel \in \mathbb{R}^d$. Let $\vc{a} \in \{0,1\}^d$ be a boolean mask.  The segmented PD threat is given by,
\begin{alignat*}{3}
    d_{\mathrm{PD}-{\rm S}}(\x, \vc{a}, \bdel) \coloneqq&~~ \underset{\vc{u} \;\in\; \mathcal{U}(\x)}{\max} ~~ \frac{1}{g_{\beta}(\x, \vc{u})} \max \left(\langle \bdel[\vc{a}], \vc{u}[\vc{a}]\rangle, 0\right). %
\end{alignat*}
\end{definition}
Here the $\vc{u}[\vc{a}]$ denotes a sub-vector whose indices are selected by the boolean mask $\vc{a}$. %
The threat model $(d_{\rm{PD-S}},\varepsilon)$ is aligned with \citet{xiao2021noise}'s proposal to learn models robust to adversarially chosen backgrounds. For each image, we generate automatic masks by prompting SAM\footnote{In particular, we use the \href{https://github.com/facebookresearch/segment-anything}{ViT-H SAM} checkpoint.} with the center pixel coordinates. \Cref{fig:PD-S} illustrates the difference between the standard PD threats and the segmented PD-S threats on 2 corruptions that are background-only and foreground-only, respectively. PD-S threat is oblivious to background corruption but more sensitive to foreground corruption. %

The quality of the PD-S threat model depends on the quality of an automatic mask generator. We note that $\ell_p$ threat models are also capable of integrating segmentation masks as they are pixel-based, but it is unclear if perceptual distance metrics can account for pixel annotation.

\subsection{Concept-aware Threat Specification}
\begin{figure}[H]
    \centering
    \includegraphics[width=0.5\linewidth, trim={1cm 0cm 0.05cm 0},clip]{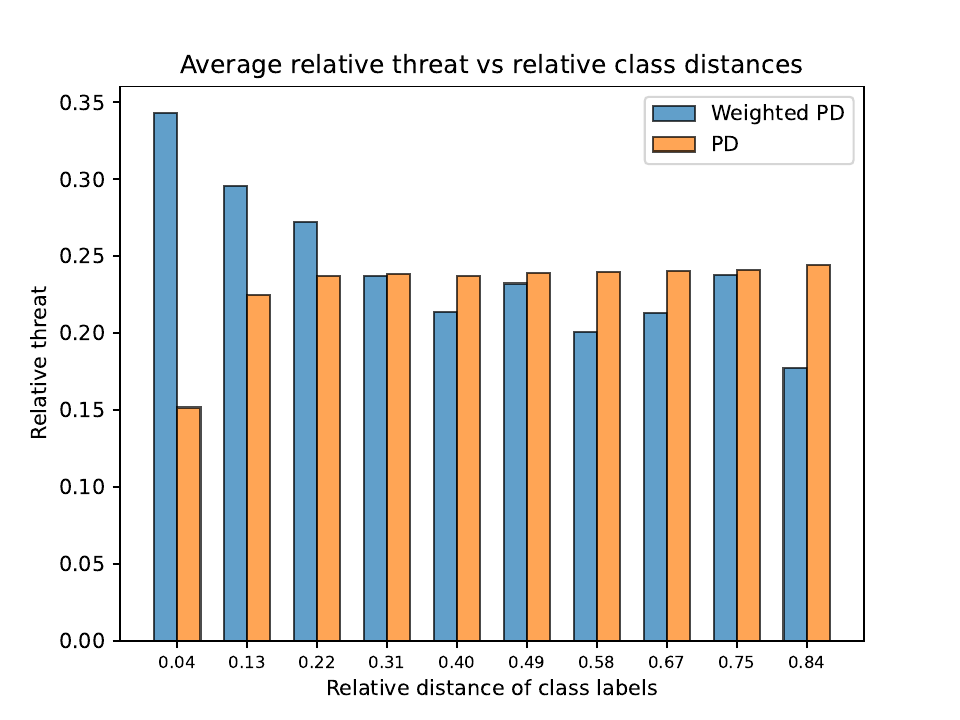}
    \caption{Relative threat vs relative inter-class distances.}
    \label{fig:weightedPD}
\end{figure}

The PD threat assumes no hierarchy of class labels; i.e.  all labels are distinct to the same degree. However, classification tasks are often accompanied with an additional hierarchy that identifies groups of class labels based on similar semantic content. We now show how to account for such a hierarchy by simply refining the choice of normalization. %
Suppose $\x$ has label $y$ and $\vc{u} = \frac{\xtil-\x}{\norm{\xtil-\x}_2}$ where $\xtil \in \samp_{c,k}$ has label c. Let $W: [C]\times [C] \in [0,1]$ denote a relative\footnote{\textit{Relative} since we require a normalized value in [0,1].} distance between class labels based on Wordnet-hierarchy (refer to the appendix \ref{app:subsec:PDW} for explicit details). We propose the weighted normalization $g_{\beta, W}(\x,\vc{u}) := \beta \cdot W(y,c)\norm{\xtil-\x}_2$, so that %
perturbations between nearby classes are weighted higher in threat. %
\begin{definition}[PD-W threat]\label{def:noniso-threat-k-weight}
Let $\x \in \cX$ with label $y$ and let $\bdel \in \mathbb{R}^d$. The weighted PD threat is defined as,
\begin{alignat*}{3}
    d_{\mathrm{PD-W}}(\x, \bdel) \coloneqq&~~ \underset{\vc{u} \;\in\; \mathcal{U}(\x)}{\max} ~~ \frac{1}{g_{\beta, W}(\x, \vc{u})} \max\left(\langle \bdel, \vc{u}\rangle, 0\right). %
\end{alignat*}
\end{definition}
The threat model $(d_{\rm PD-W}, \varepsilon)$ is weaker than $(d_{\rm PD}, \varepsilon)$, since $\mathcal{S}(\x, d_{\rm PD-W}, \varepsilon) \subseteq \mathcal{S}(\x, d_{\rm PD}, \varepsilon)$. Thus PD-W is a relaxation of the PD threat with a softer requirement of stability between nearby classes. \Cref{fig:weightedPD} depicts the average relative\footnote{Due to different scaling of PD-W and PD, we visualize average threat in each relative to maximum value on data.} threat vs relative distance $W(\cdot,\cdot)$ of class labels on the Imagenet-1k validation dataset. The decreasing PD-W threat for increasing class distance is aligned with the following natural intuition - classifiers that fail to distinguish between semantically distant classes such as \textsc{English foxhound} and \textsc{Fire truck} should have lower robust accuracy under the threat model in comparison to classifiers that fail to distinguish between similar classes such as \textsc{English foxhound} and \textsc{Irish Wolfhound}. 

In summary our novel threat specification framework enables the evaluation of robustness to adversarial perturbations and common corruptions while allowing for sensitivity to semantic regions of an image and class hierarchies by incorporating additional task annotations. As a consequence, our threat model unifies adversarial robustness with variants of corruption robustness suggested by several independent benchmarks \citep{hendrycks2019robustness, NEURIPS2021_1d497805, xiao2021noise}. 

\section{Evaluation of benchmark $\ell_p$ robust models}
Finally, we evaluate state-of-the-art robust models against our proposed threat models. Note that, as per \Cref{thm:1-rob}, one should hope for classifiers that are 1-robust to corruptions under the PD threat. Thus, the corresponding permissible set of corruptions cover regions IV and I in \Cref{fig:threat-comparison}. We evaluate robustness to corruptions in the set $\mathcal{S}(\x, d_{\rm PD}, \varepsilon) \cap \mathcal{S}(\x, d_{\infty}, \varepsilon)$, and denote the corresponding threat model $(d_{\infty} \bigcap d_{\rm PD},\; \varepsilon)$. A robust evaluation for such a threat model can be executed in practice by generating adversarial perturbations with standard \textit{AutoAttack} \citep{10.5555/3524938.3525144} for $(d_{\infty}, \varepsilon)$ followed by projection onto $\mathcal{S}(\x, d_{\rm PD}, \varepsilon)$. \Cref{tab:robust-accuracies-imagenet} shows the robustness of benchmark classifiers registered at RobustBench ordered by $d_{\infty}$ robustness (column 1) when evaluated against our $(d_{\infty} \bigcap d_{\rm PD},\; 16/255)$ threat model for the Imagenet-1k classification task.

\begin{table}[h]
    \centering
    \begin{tabular}{@{}c@{\;}x{8mm}c@{\;}c@{\;}c@{\;}c@{\;}c@{}}
    \toprule
    Benchmark Model Clean Acc. & $d_{\infty}$ & PD &  PD-S & PD-W \\ 
    \midrule
    ConvNeXt-L \citep{liu2023comprehensivestudyrobustnessimage} & 0.25  & 0.32 &0.31 &  0.31 \\ 
    Swin-B \citep{liu2023comprehensivestudyrobustnessimage}  & 0.12 & 0.21& 0.20& 0.21  \\ 
    ConvNeXt-B \citep{liu2023comprehensivestudyrobustnessimage} & 0.1  & 0.19 & 0.18& 0.18 \\  
    ConvNeXt-B-ConvStem \citep{singh2023revisitingadversarialtrainingimagenet} &   0.09  & 0.19 & 0.19& 0.19 \\     
    ViT-S-ConvStem \citep{singh2023revisitingadversarialtrainingimagenet}  & 0.05  & 0.13 & 0.12 & 0.12\\ 
    XCiT-L12 \citep{debenedetti2023lightrecipetrainrobust} & 0.04  & 0.17 & 0.18& 0.18 \\
    \bottomrule
    \end{tabular}
\caption{Evaluation of RobustBench classifiers for Imagenet-1k ($\varepsilon = 16/255$).}
\label{tab:robust-accuracies-imagenet}
\end{table}
We note that robust accuracies uniformly improve under our threat specification in comparison to ($d_{\infty}, \varepsilon$), indicating a meaningful restriction of unsafe corruptions. The ordering of robust accuracies highlights diversity, e.g. \cite{debenedetti2023lightrecipetrainrobust}'s XCiT-L12 fares better under PD threat than suggested by the $d_\infty$ leaderboard. 
The indistinguishability across PD, PD-S and PD-w 
is surprising. Similar levels of robustness w.r.t PD-W and PD indicate that models are unable to distinguish distant classes, while similar levels under PD and PD-S indicate adversarial attacks are often aligned with foreground masks. We note that adversarial attacks adapted to the threat specification often lead to a more accurate pessimistic evaluation of robustness. 

\section{Conclusion}
This work proposes a novel task-dependent threat specification, \textit{Projected Displacement} threat, that is adapted to the shape of decision boundaries based on observed training data. Unlike standard $\ell_p$-threat models, the proposed threat  
exhibits anisotropy and locality and is thus able to distinguish between safe and unsafe perturbations. The proposed threat specification framework is flexible and can effectively account for additional task information, such as image regions or label hierarchies. We invite the community to suggest further adaptive attacks and effective training strategies to encourage robustness in PD threat. %

{
	\section*{Acknowledgements}
    This research was supported by National Science Foundation (grants 2212457, 2239787 and 2031985), and the Simons Foundation (grant 814201).
    
    \bibliographystyle{ieeenat_fullname}
    \bibliography{references}

\begin{thebibliography}{60}
\providecommand{\natexlab}[1]{#1}
\providecommand{\url}[1]{\texttt{#1}}
\expandafter\ifx\csname urlstyle\endcsname\relax
  \providecommand{\doi}[1]{doi: #1}\else
  \providecommand{\doi}{doi: \begingroup \urlstyle{rm}\Url}\fi

\bibitem[Akhtar et~al.(2021)Akhtar, Mian, Kardan, and Shah]{akhtar2021advances}
Naveed Akhtar, Ajmal Mian, Navid Kardan, and Mubarak Shah.
\newblock Advances in adversarial attacks and defenses in computer vision: A
  survey.
\newblock \emph{IEEE Access}, 9:\penalty0 155161--155196, 2021.

\bibitem[Awasthi et~al.(2023)Awasthi, Mao, Mohri, and
  Zhong]{pmlr-v206-awasthi23c}
Pranjal Awasthi, Anqi Mao, Mehryar Mohri, and Yutao Zhong.
\newblock Theoretically grounded loss functions and algorithms for adversarial
  robustness.
\newblock In \emph{Proceedings of The 26th International Conference on
  Artificial Intelligence and Statistics}, pages 10077--10094. PMLR, 2023.

\bibitem[Bai et~al.(2024)Bai, Anderson, Kim, and
  Sojoudi]{bai2024improvingaccuracyrobustnesstradeoffclassifiers}
Yatong Bai, Brendon~G. Anderson, Aerin Kim, and Somayeh Sojoudi.
\newblock Improving the accuracy-robustness trade-off of classifiers via
  adaptive smoothing, 2024.

\bibitem[Bartoldson et~al.(2024)Bartoldson, Diffenderfer, Parasyris, and
  Kailkhura]{pmlr-v235-bartoldson24a}
Brian~R. Bartoldson, James Diffenderfer, Konstantinos Parasyris, and Bhavya
  Kailkhura.
\newblock Adversarial robustness limits via scaling-law and human-alignment
  studies.
\newblock In \emph{Proceedings of the 41st International Conference on Machine
  Learning}, pages 3046--3072. PMLR, 2024.

\bibitem[Cao et~al.(2019)Cao, Xiao, Cyr, Zhou, Park, Rampazzi, Chen, Fu, and
  Mao]{Cao2019AdversarialSA}
Yulong Cao, Chaowei Xiao, Benjamin Cyr, Yimeng Zhou, Wonseok Park, Sara
  Rampazzi, Qi~Alfred Chen, Kevin Fu, and Z.~Morley Mao.
\newblock Adversarial sensor attack on lidar-based perception in autonomous
  driving.
\newblock \emph{Proceedings of the 2019 ACM SIGSAC Conference on Computer and
  Communications Security}, 2019.

\bibitem[Carlini and Wagner(2017)]{Carlini2017AdversarialEA}
Nicholas Carlini and David~A. Wagner.
\newblock Adversarial examples are not easily detected: Bypassing ten detection
  methods.
\newblock \emph{Proceedings of the 10th ACM Workshop on Artificial Intelligence
  and Security}, 2017.

\bibitem[Chen et~al.(2023)Chen, Gao, Zhao, Ye, and Xu]{Chen_2023_ICCV}
Xinquan Chen, Xitong Gao, Juanjuan Zhao, Kejiang Ye, and Cheng-Zhong Xu.
\newblock Advdiffuser: Natural adversarial example synthesis with diffusion
  models.
\newblock In \emph{Proceedings of the IEEE/CVF International Conference on
  Computer Vision (ICCV)}, pages 4562--4572, 2023.

\bibitem[Croce and Hein(2020)]{10.5555/3524938.3525144}
Francesco Croce and Matthias Hein.
\newblock Reliable evaluation of adversarial robustness with an ensemble of
  diverse parameter-free attacks.
\newblock In \emph{Proceedings of the 37th International Conference on Machine
  Learning}. JMLR.org, 2020.

\bibitem[Croce et~al.(2021)Croce, Andriushchenko, Sehwag, Debenedetti,
  Flammarion, Chiang, Mittal, and Hein]{croce2021robustbench}
Francesco Croce, Maksym Andriushchenko, Vikash Sehwag, Edoardo Debenedetti,
  Nicolas Flammarion, Mung Chiang, Prateek Mittal, and Matthias Hein.
\newblock Robustbench: a standardized adversarial robustness benchmark.
\newblock In \emph{Thirty-fifth Conference on Neural Information Processing
  Systems Datasets and Benchmarks Track}, 2021.

\bibitem[Cui et~al.(2023)Cui, Tian, Zhong, Qi, Yu, and
  Zhang]{cui2023decoupledkullbackleiblerdivergenceloss}
Jiequan Cui, Zhuotao Tian, Zhisheng Zhong, Xiaojuan Qi, Bei Yu, and Hanwang
  Zhang.
\newblock Decoupled kullback-leibler divergence loss, 2023.

\bibitem[Debenedetti et~al.(2023)Debenedetti, Sehwag, and
  Mittal]{debenedetti2023lightrecipetrainrobust}
Edoardo Debenedetti, Vikash Sehwag, and Prateek Mittal.
\newblock A light recipe to train robust vision transformers, 2023.

\bibitem[Debenedetti et~al.(2024)Debenedetti, Wan, Andriushchenko, Sehwag,
  Bhardwaj, and Kailkhura]{debenedetti2024scaling}
Edoardo Debenedetti, Zishen Wan, Maksym Andriushchenko, Vikash Sehwag, Kshitij
  Bhardwaj, and Bhavya Kailkhura.
\newblock Scaling compute is not all you need for adversarial robustness.
\newblock In \emph{ICLR 2024 Workshop on Reliable and Responsible Foundation
  Models}, 2024.

\bibitem[Dodge and Karam(2017)]{dodge2017study}
Samuel Dodge and Lina Karam.
\newblock A study and comparison of human and deep learning recognition
  performance under visual distortions.
\newblock In \emph{2017 26th international conference on computer communication
  and networks (ICCCN)}, pages 1--7. IEEE, 2017.

\bibitem[Dohmatob(2019)]{Dohmatob2019GeneralizedNF}
Elvis Dohmatob.
\newblock Generalized no free lunch theorem for adversarial robustness.
\newblock In \emph{ICML}, 2019.

\bibitem[Erdemir et~al.(2021)Erdemir, Bickford, Melis, and
  Aydore]{erdemir2021adversarialrobustnessnonuniformperturbations}
Ecenaz Erdemir, Jeffrey Bickford, Luca Melis, and Sergul Aydore.
\newblock Adversarial robustness with non-uniform perturbations, 2021.

\bibitem[Fu et~al.(2024)Fu, Tamir, Sundaram, Chai, Zhang, Dekel, and
  Isola]{fu2024dreamsim}
Stephanie Fu, Netanel Tamir, Shobhita Sundaram, Lucy Chai, Richard Zhang, Tali
  Dekel, and Phillip Isola.
\newblock Dreamsim: Learning new dimensions of human visual similarity using
  synthetic data.
\newblock \emph{Advances in Neural Information Processing Systems}, 36, 2024.

\bibitem[Geirhos et~al.(2018)Geirhos, Temme, Rauber, Sch\"{u}tt, Bethge, and
  Wichmann]{NEURIPS2018_0937fb58}
Robert Geirhos, Carlos R.~M. Temme, Jonas Rauber, Heiko~H. Sch\"{u}tt, Matthias
  Bethge, and Felix~A. Wichmann.
\newblock Generalisation in humans and deep neural networks.
\newblock In \emph{Advances in Neural Information Processing Systems}. Curran
  Associates, Inc., 2018.

\bibitem[Ghazanfari et~al.(2023{\natexlab{a}})Ghazanfari, Araujo,
  Krishnamurthy, Khorrami, and Garg]{Ghazanfari2023LipSimAP}
Sara Ghazanfari, Alexandre Araujo, Prashanth Krishnamurthy, Farshad Khorrami,
  and Siddharth Garg.
\newblock Lipsim: A provably robust perceptual similarity metric.
\newblock \emph{ArXiv}, abs/2310.18274, 2023{\natexlab{a}}.

\bibitem[Ghazanfari et~al.(2023{\natexlab{b}})Ghazanfari, Garg, Krishnamurthy,
  Khorrami, and Araujo]{ghazanfari2023rlpips}
Sara Ghazanfari, Siddharth Garg, Prashanth Krishnamurthy, Farshad Khorrami, and
  Alexandre Araujo.
\newblock R-{LPIPS}: An adversarially robust perceptual similarity metric.
\newblock In \emph{The Second Workshop on New Frontiers in Adversarial Machine
  Learning}, 2023{\natexlab{b}}.

\bibitem[Ghildyal and Liu(2023)]{Ghildyal2023AttackingPS}
Abhijay Ghildyal and Feng Liu.
\newblock Attacking perceptual similarity metrics.
\newblock \emph{Trans. Mach. Learn. Res.}, 2023, 2023.

\bibitem[Gilmer et~al.(2018)Gilmer, Adams, Goodfellow, Andersen, and
  Dahl]{gilmer2018motivatingrulesgameadversarial}
Justin Gilmer, Ryan~P. Adams, Ian Goodfellow, David Andersen, and George~E.
  Dahl.
\newblock Motivating the rules of the game for adversarial example research,
  2018.

\bibitem[Gonzalez(1985)]{GONZALEZ1985293}
Teofilo~F. Gonzalez.
\newblock Clustering to minimize the maximum intercluster distance.
\newblock \emph{Theoretical Computer Science}, 38:\penalty0 293--306, 1985.

\bibitem[Goodfellow et~al.(2015)Goodfellow, Shlens, and
  Szegedy]{Goodfellow2015ExplainingAH}
Ian~J. Goodfellow, Jonathon Shlens, and Christian Szegedy.
\newblock Explaining and harnessing adversarial examples.
\newblock \emph{CoRR}, abs/1412.6572, 2015.

\bibitem[Gowal et~al.(2021{\natexlab{a}})Gowal, Qin, Uesato, Mann, and
  Kohli]{gowal2021uncoveringlimitsadversarialtraining}
Sven Gowal, Chongli Qin, Jonathan Uesato, Timothy Mann, and Pushmeet Kohli.
\newblock Uncovering the limits of adversarial training against norm-bounded
  adversarial examples, 2021{\natexlab{a}}.

\bibitem[Gowal et~al.(2021{\natexlab{b}})Gowal, Rebuffi, Wiles, Stimberg,
  Calian, and Mann]{gowal2021improvingrobustnessusinggenerated}
Sven Gowal, Sylvestre-Alvise Rebuffi, Olivia Wiles, Florian Stimberg,
  Dan~Andrei Calian, and Timothy Mann.
\newblock Improving robustness using generated data, 2021{\natexlab{b}}.

\bibitem[Hendrycks and Dietterich(2019)]{hendrycks2019robustness}
Dan Hendrycks and Thomas Dietterich.
\newblock Benchmarking neural network robustness to common corruptions and
  perturbations.
\newblock \emph{Proceedings of the International Conference on Learning
  Representations}, 2019.

\bibitem[Jacobsen et~al.(2019)Jacobsen, Behrmann, Zemel, and
  Bethge]{jacobsen2018excessive}
Joern-Henrik Jacobsen, Jens Behrmann, Richard Zemel, and Matthias Bethge.
\newblock Excessive invariance causes adversarial vulnerability.
\newblock In \emph{International Conference on Learning Representations}, 2019.

\bibitem[Kanbak et~al.(2018)Kanbak, Moosavi-Dezfooli, and Frossard]{8578565}
Can Kanbak, Seyed-Mohsen Moosavi-Dezfooli, and Pascal Frossard.
\newblock { Geometric Robustness of Deep Networks: Analysis and Improvement }.
\newblock In \emph{2018 IEEE/CVF Conference on Computer Vision and Pattern
  Recognition (CVPR)}, pages 4441--4449, Los Alamitos, CA, USA, 2018. IEEE
  Computer Society.

\bibitem[Kar et~al.(2022)Kar, Yeo, Atanov, and Zamir]{kar20223d}
O{\u{g}}uzhan~Fatih Kar, Teresa Yeo, Andrei Atanov, and Amir Zamir.
\newblock 3d common corruptions and data augmentation.
\newblock In \emph{Proceedings of the IEEE/CVF Conference on Computer Vision
  and Pattern Recognition}, pages 18963--18974, 2022.

\bibitem[Kettunen et~al.(2019)Kettunen, Härkönen, and
  Lehtinen]{kettunen2019elpipsrobustperceptualimage}
Markus Kettunen, Erik Härkönen, and Jaakko Lehtinen.
\newblock E-lpips: Robust perceptual image similarity via random transformation
  ensembles, 2019.

\bibitem[Kirillov et~al.(2023)Kirillov, Mintun, Ravi, Mao, Rolland, Gustafson,
  Xiao, Whitehead, Berg, Lo, Doll{\'a}r, and Girshick]{kirillov2023segany}
Alexander Kirillov, Eric Mintun, Nikhila Ravi, Hanzi Mao, Chloe Rolland, Laura
  Gustafson, Tete Xiao, Spencer Whitehead, Alexander~C. Berg, Wan-Yen Lo, Piotr
  Doll{\'a}r, and Ross Girshick.
\newblock Segment anything.
\newblock \emph{arXiv:2304.02643}, 2023.

\bibitem[Kurakin et~al.(2017)Kurakin, Goodfellow, and
  Bengio]{Kurakin2017AdversarialEI}
Alexey Kurakin, Ian~J. Goodfellow, and Samy Bengio.
\newblock Adversarial examples in the physical world.
\newblock \emph{ArXiv}, abs/1607.02533, 2017.

\bibitem[Laidlaw et~al.(2021)Laidlaw, Singla, and
  Feizi]{laidlaw2021perceptualadversarialrobustnessdefense}
Cassidy Laidlaw, Sahil Singla, and Soheil Feizi.
\newblock Perceptual adversarial robustness: Defense against unseen threat
  models, 2021.

\bibitem[Liu et~al.(2023)Liu, Dong, Xiang, Yang, Su, Zhu, Chen, He, Xue, and
  Zheng]{liu2023comprehensivestudyrobustnessimage}
Chang Liu, Yinpeng Dong, Wenzhao Xiang, Xiao Yang, Hang Su, Jun Zhu, Yuefeng
  Chen, Yuan He, Hui Xue, and Shibao Zheng.
\newblock A comprehensive study on robustness of image classification models:
  Benchmarking and rethinking, 2023.

\bibitem[Luo et~al.(2022)Luo, Lin, Xie, Wu, Xie, and Shen]{Luo_2022_CVPR}
Cheng Luo, Qinliang Lin, Weicheng Xie, Bizhu Wu, Jinheng Xie, and Linlin Shen.
\newblock Frequency-driven imperceptible adversarial attack on semantic
  similarity.
\newblock In \emph{Proceedings of the IEEE/CVF Conference on Computer Vision
  and Pattern Recognition (CVPR)}, pages 15315--15324, 2022.

\bibitem[Madry et~al.(2018)Madry, Makelov, Schmidt, Tsipras, and
  Vladu]{Madry2018TowardsDL}
Aleksander Madry, Aleksandar Makelov, Ludwig Schmidt, Dimitris Tsipras, and
  Adrian Vladu.
\newblock Towards deep learning models resistant to adversarial attacks.
\newblock \emph{ArXiv}, abs/1706.06083, 2018.

\bibitem[Mantiuk et~al.(2011)Mantiuk, Kim, Rempel, and
  Heidrich]{Mantiuk2011HDRVDP2AC}
Rafał~K. Mantiuk, Kil~Joong Kim, Allan~G. Rempel, and Wolfgang Heidrich.
\newblock Hdr-vdp-2: a calibrated visual metric for visibility and quality
  predictions in all luminance conditions.
\newblock \emph{ACM SIGGRAPH 2011 papers}, 2011.

\bibitem[Mintun et~al.(2021)Mintun, Kirillov, and Xie]{NEURIPS2021_1d497805}
Eric Mintun, Alexander Kirillov, and Saining Xie.
\newblock On interaction between augmentations and corruptions in natural
  corruption robustness.
\newblock In \emph{Advances in Neural Information Processing Systems}, pages
  3571--3583. Curran Associates, Inc., 2021.

\bibitem[Papernot et~al.(2016)Papernot, McDaniel, Jha, Fredrikson, Celik, and
  Swami]{papernot2016limitations}
Nicolas Papernot, Patrick McDaniel, Somesh Jha, Matt Fredrikson, Z~Berkay
  Celik, and Ananthram Swami.
\newblock The limitations of deep learning in adversarial settings.
\newblock In \emph{2016 IEEE European symposium on security and privacy
  (EuroS\&P)}, pages 372--387. IEEE, 2016.

\bibitem[Qin et~al.(2019)Qin, Carlini, Cottrell, Goodfellow, and
  Raffel]{qin2019imperceptible}
Yao Qin, Nicholas Carlini, Garrison Cottrell, Ian Goodfellow, and Colin Raffel.
\newblock Imperceptible, robust, and targeted adversarial examples for
  automatic speech recognition.
\newblock In \emph{International conference on machine learning}, pages
  5231--5240. PMLR, 2019.

\bibitem[Ravi et~al.(2024)Ravi, Gabeur, Hu, Hu, Ryali, Ma, Khedr, R{\"a}dle,
  Rolland, Gustafson, Mintun, Pan, Alwala, Carion, Wu, Girshick, Doll{\'a}r,
  and Feichtenhofer]{ravi2024sam2}
Nikhila Ravi, Valentin Gabeur, Yuan-Ting Hu, Ronghang Hu, Chaitanya Ryali,
  Tengyu Ma, Haitham Khedr, Roman R{\"a}dle, Chloe Rolland, Laura Gustafson,
  Eric Mintun, Junting Pan, Kalyan~Vasudev Alwala, Nicolas Carion, Chao-Yuan
  Wu, Ross Girshick, Piotr Doll{\'a}r, and Christoph Feichtenhofer.
\newblock Sam 2: Segment anything in images and videos.
\newblock \emph{arXiv preprint arXiv:2408.00714}, 2024.

\bibitem[Sen et~al.(2020)Sen, Zhu, Marshall, and
  Nowak]{10.1007/978-3-030-64793-3_10}
Ayon Sen, Xiaojin Zhu, Erin Marshall, and Robert Nowak.
\newblock Popular imperceptibility measures in visual adversarial attacks are
  far from human perception.
\newblock In \emph{Decision and Game Theory for Security}, pages 188--199,
  Cham, 2020. Springer International Publishing.

\bibitem[Sharif et~al.(2018)Sharif, Bauer, and Reiter]{Sharif2018OnTS}
Mahmood Sharif, Lujo Bauer, and Michael~K. Reiter.
\newblock On the suitability of lp-norms for creating and preventing
  adversarial examples.
\newblock \emph{2018 IEEE/CVF Conference on Computer Vision and Pattern
  Recognition Workshops (CVPRW)}, pages 1686--16868, 2018.

\bibitem[Singh et~al.(2023)Singh, Croce, and
  Hein]{singh2023revisitingadversarialtrainingimagenet}
Naman~D Singh, Francesco Croce, and Matthias Hein.
\newblock Revisiting adversarial training for imagenet: Architectures, training
  and generalization across threat models, 2023.

\bibitem[Stimberg et~al.(2023)Stimberg, Chakrabarti, Lu, Hazimeh, Stretcu,
  Qiao, Liu, Kaya, Rashtchian, Fuxman, Tek, and
  Gowal]{stimberg2023benchmarking}
Florian Stimberg, Ayan Chakrabarti, Chun-Ta Lu, Hussein Hazimeh, Otilia
  Stretcu, Wei Qiao, Yintao Liu, Merve Kaya, Cyrus Rashtchian, Ariel Fuxman,
  Mehmet~Nejat Tek, and Sven Gowal.
\newblock Benchmarking robustness to adversarial image obfuscations.
\newblock In \emph{Thirty-seventh Conference on Neural Information Processing
  Systems Datasets and Benchmarks Track}, 2023.

\bibitem[Szegedy et~al.(2014)Szegedy, Zaremba, Sutskever, Bruna, Erhan,
  Goodfellow, and Fergus]{Szegedy2014IntriguingPO}
Christian Szegedy, Wojciech Zaremba, Ilya Sutskever, Joan Bruna, D. Erhan,
  Ian~J. Goodfellow, and Rob Fergus.
\newblock Intriguing properties of neural networks.
\newblock \emph{CoRR}, abs/1312.6199, 2014.

\bibitem[Tram{\`{e}}r et~al.(2020)Tram{\`{e}}r, Behrmann, Carlini, Papernot,
  and Jacobsen]{DBLP:conf/icml/TramerBCPJ20}
Florian Tram{\`{e}}r, Jens Behrmann, Nicholas Carlini, Nicolas Papernot, and
  J{\"{o}}rn{-}Henrik Jacobsen.
\newblock Fundamental tradeoffs between invariance and sensitivity to
  adversarial perturbations.
\newblock In \emph{Proceedings of the 37th International Conference on Machine
  Learning, {ICML} 2020, 13-18 July 2020, Virtual Event}, pages 9561--9571.
  {PMLR}, 2020.

\bibitem[Tram{\`e}r et~al.(2020)Tram{\`e}r, Carlini, Brendel, and
  Madry]{Tramr2020OnAA}
Florian Tram{\`e}r, Nicholas Carlini, Wieland Brendel, and Aleksander Madry.
\newblock On adaptive attacks to adversarial example defenses.
\newblock \emph{ArXiv}, abs/2002.08347, 2020.

\bibitem[Tu et~al.(2022)Tu, Li, Yan, Ren, Chen, Liang, Bitar, Yumer, and
  Urtasun]{pmlr-v164-tu22a}
James Tu, Huichen Li, Xinchen Yan, Mengye Ren, Yun Chen, Ming Liang, Eilyan
  Bitar, Ersin Yumer, and Raquel Urtasun.
\newblock Exploring adversarial robustness of multi-sensor perception systems
  in self driving.
\newblock In \emph{Proceedings of the 5th Conference on Robot Learning}, pages
  1013--1024. PMLR, 2022.

\bibitem[Vassilev et~al.(2024)Vassilev, Oprea, Fordyce, and Andersen]{1224151}
Apostol Vassilev, Alina Oprea, Alie Fordyce, and Hyrum Andersen.
\newblock Adversarial machine learning: A taxonomy and terminology of attacks
  and mitigations, 2024.

\bibitem[Wang et~al.(2004)Wang, Bovik, Sheikh, and Simoncelli]{1284395}
Zhou Wang, A.C. Bovik, H.R. Sheikh, and E.P. Simoncelli.
\newblock Image quality assessment: from error visibility to structural
  similarity.
\newblock \emph{IEEE Transactions on Image Processing}, 13\penalty0
  (4):\penalty0 600--612, 2004.

\bibitem[Wang et~al.(2023)Wang, Pang, Du, Lin, Liu, and
  Yan]{wang2023betterdiffusionmodelsimprove}
Zekai Wang, Tianyu Pang, Chao Du, Min Lin, Weiwei Liu, and Shuicheng Yan.
\newblock Better diffusion models further improve adversarial training, 2023.

\bibitem[Wong and Kolter(2018)]{pmlr-v80-wong18a}
Eric Wong and Zico Kolter.
\newblock Provable defenses against adversarial examples via the convex outer
  adversarial polytope.
\newblock In \emph{Proceedings of the 35th International Conference on Machine
  Learning}, pages 5286--5295. PMLR, 2018.

\bibitem[Wu et~al.(2020)Wu, Lim, Davis, and
  Goldstein]{wu2020makinginvisibilitycloakreal}
Zuxuan Wu, Ser-Nam Lim, Larry Davis, and Tom Goldstein.
\newblock Making an invisibility cloak: Real world adversarial attacks on
  object detectors, 2020.

\bibitem[Xiao et~al.(2018)Xiao, Zhu, Li, He, Liu, and Song]{xiao2018spatially}
Chaowei Xiao, Jun-Yan Zhu, Bo Li, Warren He, Mingyan Liu, and Dawn Song.
\newblock Spatially transformed adversarial examples.
\newblock In \emph{International Conference on Learning Representations}, 2018.

\bibitem[Xiao et~al.(2021)Xiao, Engstrom, Ilyas, and Madry]{xiao2021noise}
Kai~Yuanqing Xiao, Logan Engstrom, Andrew Ilyas, and Aleksander Madry.
\newblock Noise or signal: The role of image backgrounds in object recognition.
\newblock In \emph{International Conference on Learning Representations}, 2021.

\bibitem[Zhang et~al.(2019)Zhang, Yu, Jiao, Xing, Ghaoui, and
  Jordan]{pmlr-v97-zhang19p}
Hongyang Zhang, Yaodong Yu, Jiantao Jiao, Eric Xing, Laurent~El Ghaoui, and
  Michael Jordan.
\newblock Theoretically principled trade-off between robustness and accuracy.
\newblock In \emph{Proceedings of the 36th International Conference on Machine
  Learning}, pages 7472--7482. PMLR, 2019.

\bibitem[Zhang et~al.(2011)Zhang, Zhang, Mou, and Zhang]{5705575}
Lin Zhang, Lei Zhang, Xuanqin Mou, and David Zhang.
\newblock Fsim: A feature similarity index for image quality assessment.
\newblock \emph{IEEE Transactions on Image Processing}, 20\penalty0
  (8):\penalty0 2378--2386, 2011.

\bibitem[Zhang et~al.(2018)Zhang, Isola, Efros, Shechtman, and Wang]{8578166}
Richard Zhang, Phillip Isola, Alexei~A. Efros, Eli Shechtman, and Oliver Wang.
\newblock The unreasonable effectiveness of deep features as a perceptual
  metric.
\newblock In \emph{2018 IEEE/CVF Conference on Computer Vision and Pattern
  Recognition (CVPR)}, pages 586--595, Los Alamitos, CA, USA, 2018. IEEE
  Computer Society.

\bibitem[Żelasko et~al.(2021)Żelasko, Joshi, Shao, Villalba, Trmal, Dehak,
  and Khudanpur]{żelasko2021adversarialattacksdefensesspeech}
Piotr Żelasko, Sonal Joshi, Yiwen Shao, Jesus Villalba, Jan Trmal, Najim
  Dehak, and Sanjeev Khudanpur.
\newblock Adversarial attacks and defenses for speech recognition systems,
  2021.

\end{thebibliography}
}

\appendix 
\renewcommand{\appendixpagename}{APPENDIX}
\begin{center}  
\appendixpagename
\end{center}

\section{Expanded Discussion}
\subsection{Proof of Theorem \ref{thm:1-rob}}\label{app:1-rob}
\textbf{Part I.} 
Suppose the true labeling function $h^*$ is not $1$-robust at an $\x\in \cX$ with $y=h^\star(\x)$. Then, applying \cref{def:eps-robust} with $\epsilon = 1$, there exists a perturbation $\bdel \in \mathcal{S}(d^*_{\rm PD},\x,1)\cap (\cX-\{\x\})$ such that $\x+\bdel \in \cup_{c\neq y} \cX_{c}$. In other words, there exists an unsafe direction $\vc{u}\in \mathcal{U}^*(\x)$ and a step size $t > 0$ such that $\bdel = t\vc{u}$. By the definition of $g^\star$, we see that $t > g^\star(\x,\vc{u})$ and hence, 
\begin{align}\label{eq:normalization-1rob}
d^*_{\mathrm{PD}}(\x, \bdel) = d^*_{\mathrm{PD}}(\x, t \vc{u}) 
\geq 
\frac{1}{g^*(\x,\vc{u})} \max\left(\langle t \vc{u}, \vc{u}\rangle, 0\right)
= \frac{t}{g^*(\x,\vc{u})} 
> 1.    
\end{align}
This is a contradiction since we assumed that $\bdel \in \mathcal{S}(d^*_{\rm PD},\x,1)$. 
Hence $h^*$ must be $1$-robust w.r.t. $d_{\mathrm{PD}}$. 

\noindent \textbf{Part II.}
	Note if $h$ is 0-robust at any input $\x$ then h misclassifies $\x$. 
    Let $\x\in \cX_{y}$ be such that $h$ is at most $\epsilon$-robust for some $0 < \epsilon < 1$. 
    Hence there exists a perturbation $\bdel \in \mathcal{S}(d^*_{\rm PD},\x,1) \cap (\cX - \{\x\})$ such that, $h(\x+\bdel) \neq h^*(\x)$. For such a perturbation $\bdel$, there are two cases. 

    \textbf{Case 1 :} $ h(\x+\bdel) \neq h^*(\x+\bdel)$.
    
    In this case, the conclusion follows since $\x+\bdel$ is misclassified by $h$.  

    \textbf{Case 2 :} $h(\x+\bdel) = h^*(\x+\bdel) =c$ for some $c\neq y$.
    
    Since $\x+\bdel \in \cX_c$. There necessarily exists an unsafe direction $\vc{u} \in \mathcal{U}^*(\x)$ such that $\bdel = t \vc{u}$ for some $t\geq 0$. Further, by definition of the normalization function $t > g^*(\x, \vc{u})$. Thus, 
    \begin{align*}
    d^*_{\rm PD}(\x, \bdel) 
    \geq
        \frac{1}{g^*(\x,\vc{u})} \max\left(\langle t \vc{u}, \vc{u}\rangle, 0\right)
        = \frac{t}{g^*(\x,\vc{u})} 
    > 1.    
    \end{align*}
    This is a contradiction since $\bdel \in \mathcal{S}(d^*_{\rm PD},\x,1) \cap (\cX - \{\x\})$. Hence this case is not possible. 
    
    To summarize if there exists an input $\x\in \cX_{y}$ such that $h$ is at most $\epsilon$-robust for some $0< \epsilon < 1$, then any perturbation $\bdel \in \mathcal{S}(d^*_{\rm PD},\x,1) \cap (\cX - \{\x\})$ such that $h(\x+\bdel) \neq h^*(\x)$ is necessarily such that $\x+\bdel$ is misclassified by $h$, i.e., $h(\x+\bdel) \neq h^*(\x+\bdel)$. Hence the conclusion follows.

\subsection{Choosing $k$-subset $\samp_{c,k}$} \label{sec:ksubset}

To obtain a representative subset of unsafe directions $\mathcal{U}(\x)$ we find a $k$-subset $\samp_{c,k}$ of $\samp_c$ by solving a clustering-type optimization problem, %
\begin{align} \label{eq:ksubset-cosinesim}
    \samp_{c,k} :=  
    \underset{|A|=k,\; A \;\subseteq \;\samp_c}{\arg\max} \; %
    f(A),
\end{align}
\text{where}
\begin{align*}
    f(A) := \min_{\x \in \samp_c} \; 
    \max_{\vc{a} \in A} \;
    \frac{\langle \x, \vc{a} \rangle}{\norm{\x}_2\norm{\vc{a}}_2}.
\end{align*}
The objective in \Cref{eq:ksubset-cosinesim} defines a discrete $k$-center problem for which we can find an approximate minimizer using the classical greedy 2-approximation algorithm \citep{GONZALEZ1985293} which starts from a randomly selected element of $\samp_c$, and greedily expands the selection $k$ times, each time adding the best element in $\samp_c$ according to \cref{eq:ksubset-cosinesim}. \Cref{alg: greedykcenter} has a computational complexity of $\mathcal{O}(k^2|\samp_c|)$. 
The quality of this greedy approximation depends on the choice of the initial element. 
Exploring other strategies to select subsets $\samp_{c,k}$ remains an interesting future direction. 
\begin{algorithm}[h]
    \centering
    \caption{Choosing $k$-subset of $\samp_{c}$: Greedy $k$-center approximation}\label{alg: greedykcenter}
    \begin{algorithmic}
        \State Sample uniformly at random $a \sim \samp_c$
        \State Initialize : $A \gets \{ a \}$
        \Repeat
        \State Find $\vc{b} \in \samp_c$ with minimal cosine similarity to any $\vc{a} \in A$,
        \[
        \vc{b} := 
        \underset{\x \in \samp_c}{\arg \min} 
        \max_{\vc{a} \in A} \;
        \frac{\langle \x, \vc{a} \rangle}{\norm{\x}_2\norm{\vc{a}}_2}
        \]
        \State $A \gets A \cup \{b\}$.
        \Until $|A| = k$. 
        \State \textbf{Return}  $A$ %
    \end{algorithmic}
\end{algorithm}

\subsection{Projection onto sub-level sets}
Each sublevel set is the intersection of halfspaces, 
\[
\mathcal{S}(\x, d_{\rm PD}, \varepsilon) = \underset{\vc{u} \in \mathcal{U}(\x)}{\cap} \{\bdel \in \mathbb{R}^d ~|~  \langle \bdel, \vc{u} \rangle \leq \varepsilon\cdot g(\x,\vc{u}) \}
\]
We let $\mathcal{C}_{\x,\vc{u}} := \{\bdel \in \mathbb{R}^d ~|~  \langle \bdel, \vc{u} \rangle \leq \varepsilon\cdot g(\x,\vc{u})\}$ denote the half space defined by the unsafe direction $\vc{u}$ at input $\x$. For each individual halfspace we denote $P_{\mathcal{C}_{\x,\vc{u}}}$ the projection operator defined as, 
\[
P_{\mathcal{C}_{\x,\vc{u}}}(\bdel) := \bdel - \max\left( \langle \bdel, \vc{u} \rangle - \varepsilon g(\x,\vc{u}) , 0\right)\vc{u}.
\]
To obtain a projection operator for the sublevel sets one can employ a greedy procedure (see \Cref{alg: greedyproject}) that alternates between projection onto the halfspaces $\mathcal{C}_{\x,\vc{u}}$. 

\begin{algorithm}[H]
    \centering
    \caption{Greedy Projection}\label{alg: greedyproject}
    \begin{algorithmic}
        \Require Nonempty closed convex sets $C_i$ for $1\leq i\leq T$. 
        \Require Input $\vc{a} \in \mathbb{R}^d$. 
        \Require Iteration hyper-parameter $N\geq1$.
        \Ensure Projection onto intersection of convex sets $\cap_i C_i$.
        \RepeatN{$N$}
        \State Select farthest convex set $C_j$ and project to $C_j$, 
        \begin{align}
        C_j &\gets \underset{C_i}{\arg\max}~ \norm{\vc{a}-P_{C_i}(\vc{a})}_2 \\
        \vc{a} &\gets P_{C_j}(\vc{a}).
        \end{align}
        \End 
        \State \textbf{Return} $\vc{a}$. 
    \end{algorithmic}
\end{algorithm}

In practice, we leverage the linearity of the PD-threat by utilizing the lazy projection algorithm $\bdel \rightarrow \frac{\varepsilon}{d_{\rm PD}(\x, \bdel)} \bdel$ in time-and compute-constrained settings.

\subsection{Additional Technical characteristics of PD}

\textbf{Scope of Design}.
We emphasize that the PD threat is designed to disentangle safe and unsafe perturbations. By construction, the unsafe directions $\mathcal{U}(\x)$ only contain perturbations that alter the class label. Thus, the PD threat is not expected to differentiate two safe perturbations. The threat of two different safe perturbations that retain the class label need not be ordered by the perceptual similarity, which might require the extraction of relevant image features at multiple levels of resolution. As such the PD threat is not a suitable replacement for neural perceptual distance metrics, and human-evaluation studies such as two-alternative forced choice (2AFC) testing are out of scope for the proposed design. In a similar vein,  unsafe directions only correspond to observations within the data domain. Along a safe direction, the  threat is not normalized to the boundaries of the data domain. Hence the PD threat (even exact threat $d^*_{\rm PD}$) is not designed to identify out-of-distribution (OOD) data.%

\begin{figure}[h]
    \centering
    \includegraphics[width=0.45\linewidth, trim={0.05cm 0 0.05cm 0},clip]{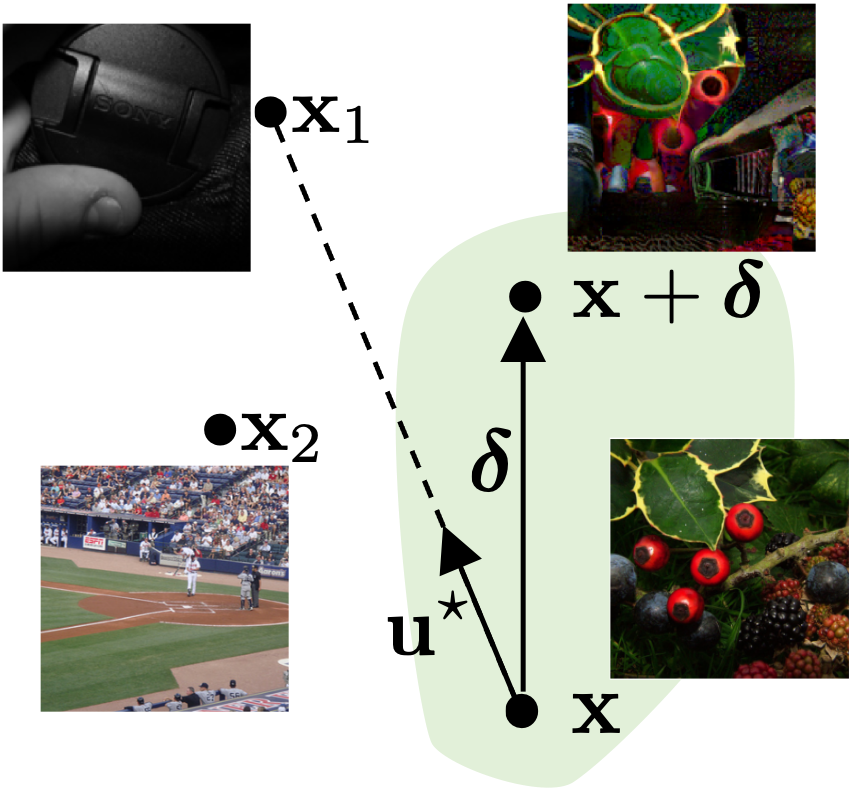}
    \caption{Attribution of threat}
    \label{fig:rosehip-adv}
\end{figure}

\textbf{Attribution}.
In \Cref{fig:rosehip-adv}, the perturbation $\bdel$ is computed via an adversarial attack on a benchmark model from \citet{liu2023comprehensivestudyrobustnessimage} at input $\x$ with label $y$. Let $u^* := \underset{\vc{u} \;\in\; \mathcal{U}(\x)}{\arg\max} ~~ \frac{1}{g_{\beta}(\x, \vc{u})} \max\left(\langle \bdel, \vc{u}\rangle, 0\right)$, be the unsafe direction most aligned with the perturbation $\bdel$. Since $u^* \in \mathcal{U}(\x)$, there exists a point $\x_1 \in \samp_{c,k}$ where $c\neq y$ such that $y^*=\frac{\x_1-\x}{\norm{\x_1-\x}_2}$. Hence for each perturbation $\bdel$ and input $\x$, the PD-threat identifies a point $\x_1$ that is most aligned with the direction of perturbation. This feature of the PD threat enables a direct attribution of the threat to observed training data points. 

\textbf{Bottom-up Perspective}.
We first note that the threat function at each input $\x$, can be re-arranged as a particular large-width single layer feed-forward neural network $h_{\mathrm{PD},\x}$. Here the neural networks $h_{\mathrm{PD},\x}$ has parameters $\{\samp_{c,k}\}_{c \in [C]}$. On the Imagenet dataset with $C=1,000$ labels, and $k=50$, this amounts to a neural network with $\approx 7.5$ billion parameters. However, unlike the pretraining required to compute vision-language embeddings, the PD threat describes a randomized neural network based on observed data that does not require an iterative gradient based learning of parameters. The only computation needed is the selection of representative subset $\samp_{c,k}$ for each label $c$. Thus the PD-threat can be viewed as a bottom-up definition of a neural embedding that requires no training and is instead motivated as the heuristic empirical approximation of a principled exact non-parametric threat function $d^*_{\rm PD}$.

\subsection{Weighted Threat Specification}\label{app:subsec:PDW}
\Cref{def:noniso-threat-k-weight} requires a relative distance between class labels $W : [C] \times [C] \rightarrow [0,1]$. We note that $W(y,c)$ is the weight assigned to threats where $y$ is the class label of the original input under perturbation and $c$ is the class label associated with the unsafe direction. The role of the arguments are distinct. In this section we identify 3 distinct approaches to computing such a relative distance between class labels. For each approach, the relative weights are computed by scaling w.r.t minimum and maximum for any fixed class $y$ and varying class $c'$ of the unsafe directions. This additional normalization ensures comparability of weights across different approaches. The final weights $W(y,c)$ used to define the weighted threat specification combines all 3 approaches. 

\begin{definition}[Euclidean Relative Weights]
We define the class distance based on Euclidean norm as the average $\ell_2$ distance between the selected subsets\footnote{Recall, $\samp_{c,k}$ is the representative subset chosen to formulate the threat specification.} of training data $\{\samp_{c,k}\}_{c=1}^C$, 
\[
\mathrm{L2}(y,c) := \expect_{\substack{\x \sim \mathrm{Unif}(\samp_{y,k}), \\ \xtil \sim \mathrm{Unif}(\samp_{c,k})}} \norm{\x-\xtil}_2,
\]
where $\mathrm{Unif}(\cdot)$ denotes sampling uniformly at random. The relative class distance $W_{\rm Euclidean} : [C] \times [C] \rightarrow [0,1]$ is the defined as,
\[
W_{\rm Euclidean}(y,c) := \frac{
\mathrm{L2}(y,c) - \min_{c_1} \mathrm{L2}(y,c_1)
}{
\max_{c_2} \mathrm{L2}(y,c_2) - \min_{c_1} \mathrm{L2}(y,c_1)
}.
\] 
\end{definition}
As explained in the motivation, $\ell_p$ norms are insufficient to measure perceptual similarity between images of distinct class labels. Hence we propose to account for perceptual similarity using DreamSim.
\begin{definition}[DreamSim Relative Weights]
We define the class distance based on DreamSim as,
\[
\mathrm{DS}(y,c) := \expect_{\substack{\x \sim \mathrm{Unif}(\samp_{y,k}), \\ \xtil \sim \mathrm{Unif}(\samp_{c,k})}} \mathrm{DreamSim}(\x, \xtil). 
\]
The relative class distance $W_{\rm DS} : [C] \times [C] \rightarrow [0,1]$ is the defined as,
\[
W_{\rm DreamSim}(y,c) := 
\frac{
\mathrm{DS}(y,c) - \min_{c_1} \mathrm{DS}(y,c_1)
}{
\max_{c_2} \mathrm{DS}(y,c_2)
-\min_{c_1} \mathrm{DS}(y,c_1)
}
\] 
\end{definition}
DreamSim is finetuned to approximate human perceptual similarity judgements however it is unclear if DreamSim can explicitly account for the concept hierarchy of Imagenet-1k class labels provided by WordNet. For example, images of class labels \textsc{Hen} and \text{Ostrich} can be deemed perceptually distinct but are close semantically as they both correspond to bird categories. 

Next, we discuss a relative distance that explicitly accounts for semantic similarity based on class hierarchy. For Imagenet-1k class labels $[C]$, the associated WordNet hierarchy can be represented as the tree $\mathrm{WordNet}(V,E)$ where $[C] \subset V$ and 
$V$ is the set of WordNet classes and the edge set $E$ contains an edge $(v_1, v_2)$ if $v_1$ is a sub-class of $v_2$ or vice-versa. For any pair of classes $(v_1, v_2)$, the lowest common ancestor $\mathrm{LCA} : V \times V \rightarrow V$ function outputs the lowest (i.e. deepest) class $\mathrm{LCA}(v_1, v_2)$ that has both $v_1$ and $v_2$ as descendants. Let $v_{\rm root}$ denote the root class (\textsc{Entity} for Imagenet-1k) such that all classes in $V\backslash \{v_{\rm root}\}$ are descendants of $v_{\rm root}$. Let $\mathrm{d}_{v}$ denote the length of the minimal path from root $v_{\rm root}$ to class $v$. 
\begin{definition}[WordNet Relative Weights]
For any two pairs of classes $(v_1, v_2)$, we define the class distance based on the WordNet hierarchy as the length of the minimal path connecting the two classes $v_1$ and $v_2$ (through $\mathrm{LCA}(v_1, v_2)$), 
\[
\mathrm{dist}_{\rm LCA}(v_1,v_2) := d_{v_1} + d_{v_2} - 2d_{\mathrm{LCA}(v_1, v_2)}
\]
The relative class distance based on the WordNet class hierarchy, $W_{\rm WordNet} : [C] \times [C] \rightarrow [0,1]$ is defined as,
\[
W_{\rm WordNet}(y,c) := \frac{
\mathrm{dist}_{\rm LCA}(y,c) - \min_{c_{1}} 
\mathrm{dist}_{\rm LCA}(y,c_{1})}{\max_{c_2} 
\mathrm{dist}_{\rm LCA}(y,c_2) - \min_{c_{1}} 
\mathrm{dist}_{\rm LCA}(y,c_{1})}
\] 
\end{definition}

\begin{definition}[Threat Specification Relative Weights]
The relative weights $W : [C]\times [C] \rightarrow [0,1]$ used to define the weighted threat specification are explicitly defined as, 
\[
W(y,c) := %
\Big(
\min \left\{ W_{\rm Euclidean}(y,c), W_{\rm DreamSim}(y,c) , W_{\rm WordNet}(y,c) \right\}
\Big)^{2}.
\]
\end{definition}
A smaller value of $W(y,c) \approx 0$ indicates that the class labels $y$ and $c$ are \textit{nearby} by at least one of the 3 relative distances based on Euclidean norm, DreamSim or WordNet Hierarchy. 
A smaller value of $W(y,c)$ indicates that perturbations $\bdel$ on inputs $\x$ with label $y$ that are aligned with unsafe directions $\vc{u}$ of label $c$ have a larger threat $d_{\rm PD-W}(\x,\bdel)$, thus the threat of perturbations between nearby classes is amplified.

\section{Illustrative Examples}\label{appendix:illustrative}
\begin{figure}[h]
    \centering
    \centerline{\includegraphics[width=1.3\linewidth, height=\plotheight\textheight, keepaspectratio]{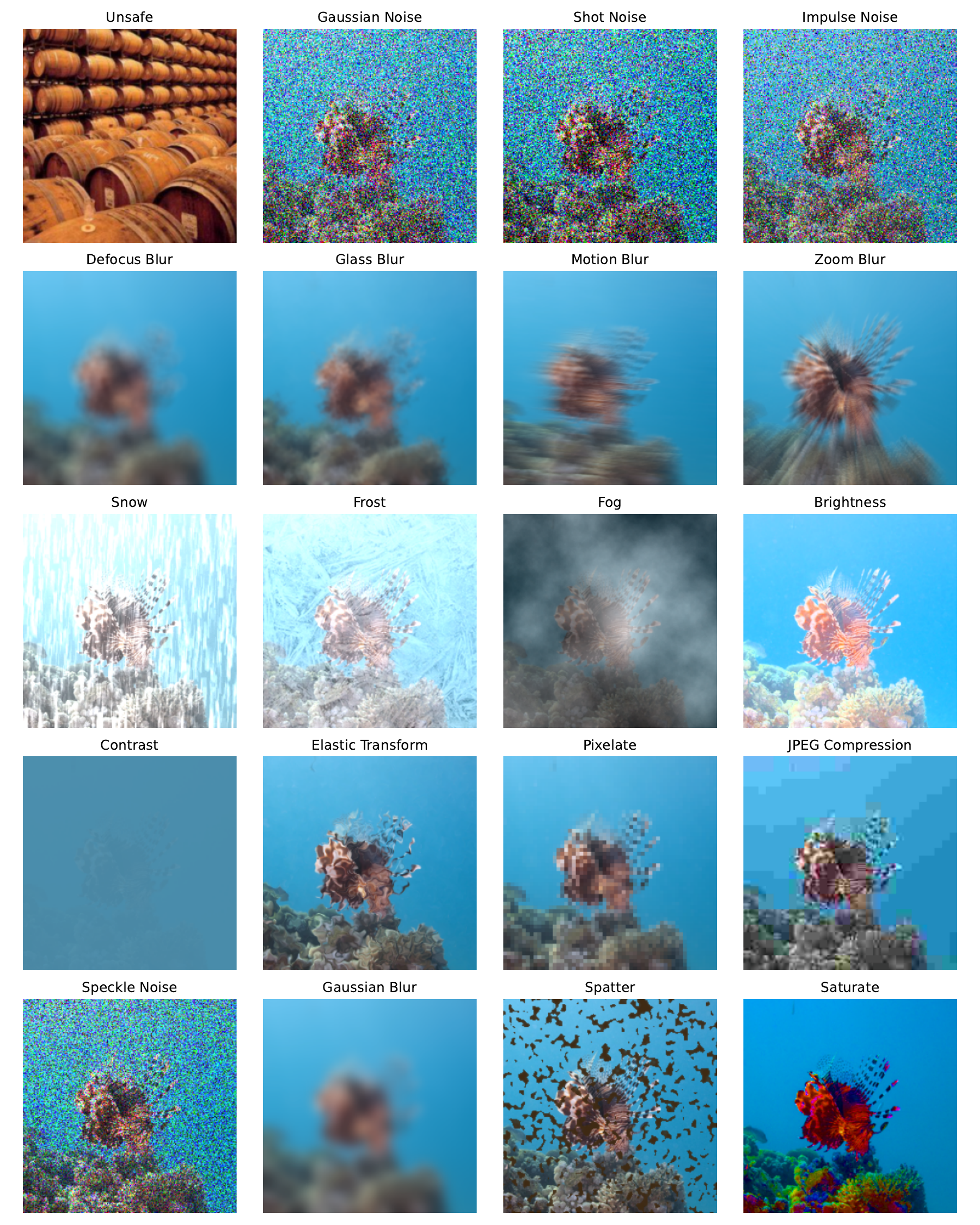}}
    \caption{Imagenet-$C$ corruptions on a sample image of class \textsc{Lionfish}.}
    \label{fig:lionfish_corruptions_c}
\end{figure}

\begin{figure}[h]
    \centering
    \centerline{\includegraphics[width=1.3\linewidth, height=\plotheight\textheight, keepaspectratio]{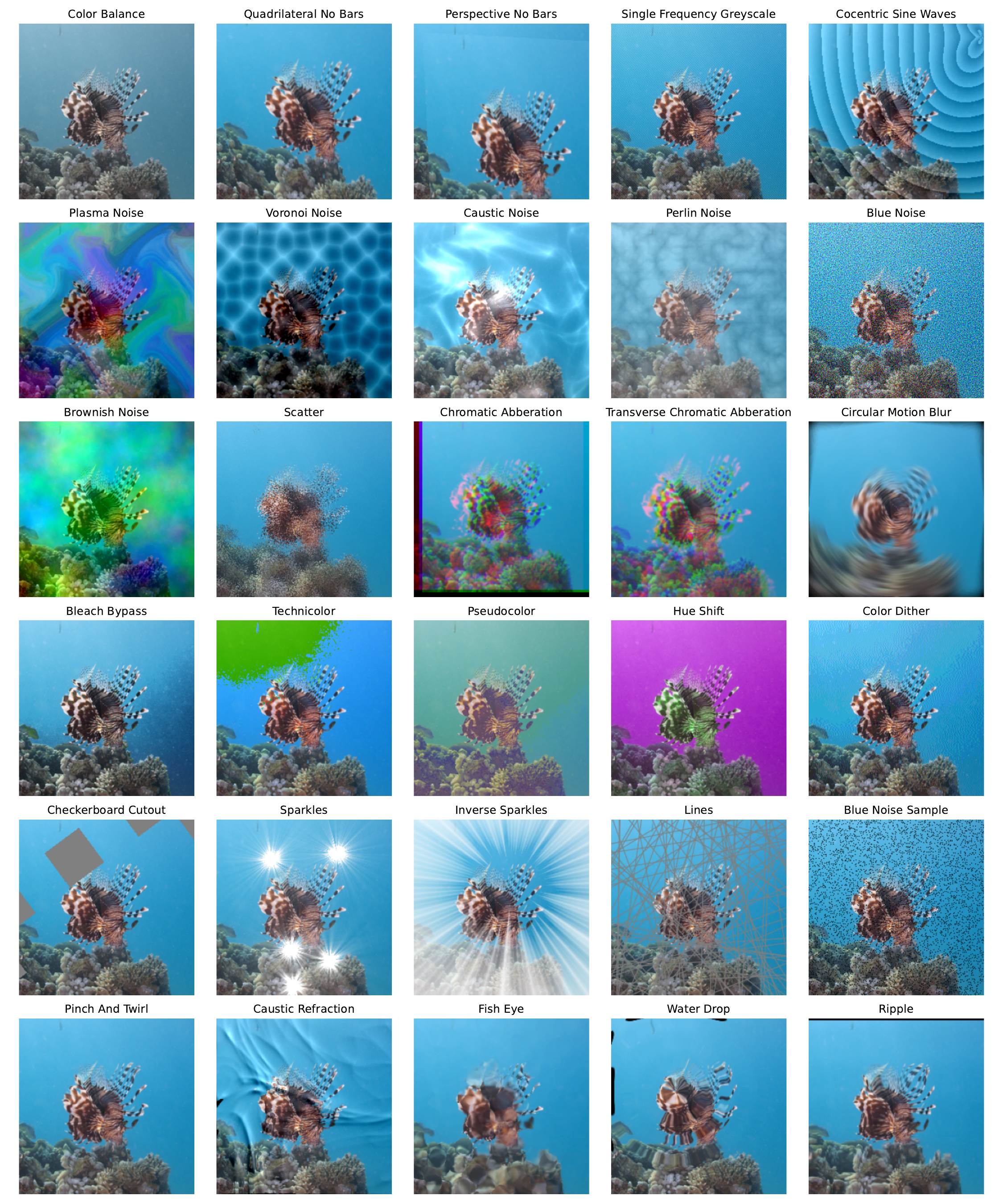}}
    \caption{Imagenet-$\bar{C}$ corruptions on a sample image of class \textsc{Lionfish}.}
    \label{fig:lionfish_corruptions_c_bar}
\end{figure}

Following \citet{NEURIPS2021_1d497805}'s protocol, each corruption style is computed in-memory to avoid additional noise incurred from compression quality of images saved to disk. 
\Cref{fig:lionfish_corruptions_c} and \Cref{fig:lionfish_corruptions_c_bar} show the corruptions of an original image $\x$ with label \textsc{Lionfish}. \Cref{tab:LionFish_c} and \Cref{tab:LionFish_c_bar} show the amount of threat assesed for each corruption by 6 threat models : (1) the $\ell_p$ threat models, $d_{\infty}$ and $d_2$, (2) the proposed PD threat models, $d_{\rm PD}$, $d_{\rm PD-W}$ and $d_{\rm PD-S}$, and finally (3) the Dreamsim threat model $d_{\rm DS}$. In \Cref{tab:LionFish_c} and \Cref{tab:LionFish_c_bar} the threats across different threat models are not comparable due to different scalings (for e.g. $d_{\infty}, d_{\rm DS}$ range between $[0,1]$) but the others are not limited to an interval. Note, the threats $d_2$ and $d_{\rm PD-W}$ are scaled by a constant factor for readability.

The corruptions are sourced from Imagenet-$C$ \cite{hendrycks2019robustness} and Imagenet-$\bar{C}$ \cite{NEURIPS2021_1d497805}. The Imagenet-$C$ corruptions are at severity $5$ (maximum 5). The Imagenet-$\bar{C}$ corruptions includes the full list of 30 corruptions at severity 5 (maximum 10). We note that the larger experiments on comparison of average threat (for eg. \Cref{fig:threat-comparison}) only include the subset of 10 Imagenet-$\bar{C}$ corruptions that are considered semantically distinct from Imagenet-$C$ \cite{NEURIPS2021_1d497805}. For any fixed threat model, the threats for different corruption styles vary. Selective entries are colored \textit{red} if the threat assessed for the corruption style is as large as the threat of the unsafe perturbation. Entries are colored \textit{orange} if the threat of the corruption style is at least half of the threat of the perturbation. The colors are meant only for illustrative purposes to highlight the ability of each threat model to separate safe and unsafe perturbations. Evidently most of the safe corruptions are rated as high threat by $d_{\infty}$. We note that both $d_2$ and $d_{\rm DS}$ rate noise corruptions as high threat but $d_{\rm PD}, d_{\rm PD-W}$ and $d_{\rm PD-S}$ do not. Weather corruptions are uniformly rated as high threat. Implementation of PD-threat and other necessary files can be found at our \href{https://github.com/ramcha24/nonisotropic}{github repository}.

\begin{table}[h]
    \centering
    \begin{tabular}{cccccccc}
    \toprule 
    \textsc{Category} & \textsc{Style} & $d_{\infty}$ & $d_{2}$ & $d_{\rm PD}$ & $d_{\rm PD-W}$ & $d_{\rm PD-S}$ & $d_{\rm DS}$ \\
    \toprule 
    {\color{red} Unsafe } 
    & {\color{red} Unsafe } & {\color{red} 0.90 } & {\color{red} 0.40  } & {\color{red} 3.30  } & {\color{red} 1.79  }& {\color{red} 2.51  }& {\color{red} 0.64  }\\
    \midrule 
    Noise
    & Gaussian Noise & {\color{orange} 0.88 } & {\color{orange} 0.28 }  & 0.51  & 0.28  & 0.49  & {\color{orange} 0.36}   \\
     & Shot Noise & {\color{orange} 0.89 } & {\color{orange} 0.30 }  & 0.61  & 0.30  & 0.46  & {\color{orange} 0.35 }   \\
     & Impulse Noise & {\color{red} 0.90 } & {\color{orange} 0.27 }  & 0.57  & 0.31  & 0.52  & {\color{orange} 0.35 }   \\ 
      & Speckle Noise & {\color{red} 0.91 } & {\color{orange} 0.24 }  & 0.46  & 0.24  & 0.35  & {\color{orange} 0.32 }  \\
     \midrule 
    Blur 
    & Defocus Blur & {\color{orange} 0.64 } & 0.08  & 0.40  & 0.15  & {\color{orange} 1.28}  & 0.23   \\
     & Glass Blur & {\color{orange} 0.59 } & 0.07  & 0.38  & 0.15  & 1.24  & 0.16  \\
     & Motion Blur & {\color{orange} 0.66 } & 0.09  & 0.43  & 0.19  & {\color{orange} 1.29 }  & 0.18  \\
     & Zoom Blur & {\color{orange} 0.51 } & 0.08 & 0.37   & 0.18  & 0.96  & 0.14  \\ 
     & Gaussian Blur & {\color{orange} 0.63 } & 0.07  & 0.42  & 0.17  & {\color{orange} 1.29 }  & 0.26  \\
     \midrule 
    Weather 
    & Snow & {\color{orange} 0.82 } & {\color{orange} 0.38 } & {\color{orange} 2.53 }  & 0.79  & {\color{orange} 2.19 } & {\color{orange} 0.53 } \\
     & Frost & {\color{orange} 0.58 } & {\color{orange} 0.34 }  & {\color{orange} 2.31}  & 0.68  & {\color{orange}2.40 }  & {\color{orange} 0.44 }  \\
     & Fog & {\color{orange} 0.61 } & {\color{orange} 0.22}  & {\color{orange}  1.87 }  & {\color{orange}  0.99 } & {\color{orange} 1.76 }  & {\color{orange} 0.49 } \\
     \midrule
    Compression 
    & Pixelate & {\color{orange} 0.54 } & 0.06  & 0.22  & 0.09  & 0.76  & 0.11  \\
     & JPEG  & {\color{orange} 0.51 } & 0.06  & 0.12  & 0.05  & 0.39  & 0.10  \\
    \midrule
   Digital 
   & Brightness & {\color{orange} 0.47 } & {\color{orange} 0.26 } & 1.41  & 0.48  & {\color{orange} 2.07 } & 0.20  \\
     & Contrast & {\color{orange} 0.60 } & 0.16  & 1.54  & 0.74  & {\color{orange} 2.13 }  & {\color{orange} 0.48 }  \\
    & Elastic Transform & {\color{orange} 0.69} & 0.16  & 0.25  & 0.11  & 0.89  & 0.01  \\
     & Saturate & {\color{orange} 0.87 } & 0.18  & 0.71  & 0.27  & 1.15  & 0.24  \\
    \midrule
    Occlusion 
    & Spatter & {\color{orange} 0.81 } & 0.16  & 0.63  & 0.31  & 0.49  & 0.25  \\
    \bottomrule
    \end{tabular}
    \caption{Threats evaluated on Imagenet-$C$ corruptions in \Cref{fig:lionfish_corruptions_c}.}
    \label{tab:LionFish_c}
\end{table}

\begin{table}[h]
    \centering
    \begin{tabular}{cccccccc}
    \toprule 
    \textsc{Category} & \textsc{Style} & $d_{\infty}$ & $d_{2}$ & $d_{\rm PD}$ & $d_{\rm PD-W}$ & $d_{\rm PD-S}$ & $d_{\rm DS}$ \\
    \toprule 
    {\color{red} Unsafe } 
    & {\color{red} Unsafe } & {\color{red} 0.90 } & {\color{red} 0.40  } & {\color{red} 3.30  } & {\color{red} 1.79  } & {\color{red} 2.51  }& {\color{red} 0.64  }\\
    \midrule 
    Noise
    & Plasma Noise & 0.29 & 0.13  & 0.62  & 0.25  & 0.72  & 0.17    \\
     & Voronoi Noise & 0.30 & 0.18  & 1.04  & 0.37  & 0.91  & 0.18    \\
     & Caustic Noise & {\color{orange} 0.59 } & 0.14  & 0.71  & 0.20  & {\color{orange}  1.31}  & 0.09   \\ 
      & Perlin Noise & 0.28 & 0.11  & 0.88  & 0.37  & 0.95  & 0.31  \\
     & Blue Noise & {\color{orange} 0.87 }& {\color{orange} 0.22 }  & 0.26  & 0.15  & 0.24  & 0.25  \\
     & Brownish Noise & {\color{orange} 0.48 } & 0.16  & 0.56  & 0.22  & 0.91  & 0.29  \\
     & Blue Noise Sample &{\color{red} 0.90 } & 0.16  & 0.44  & 0.19  & 0.30  & 0.29  \\
     \midrule 
    Blur 
     & Cocentric Sine Waves & 0.12 & 0.07  & 0.03  & 0.01  & 0.07  & 0.16  \\
     & Scatter & {\color{orange} 0.70 } & 0.10  & 0.34  & 0.14  & 1.13  & 0.12  \\
     & Chromatic Abberations & {\color{orange} 0.89} & 0.17  & 0.76  & 0.31  & {\color{orange} 1.48 }  & 0.22  \\
     & Transverse Chromatic Abberation &{\color{orange}  0.70 } & 0.10  & 0.38  & 0.18  & 0.94  & 0.20  \\ 
     & Circular Motion Blur & {\color{orange} 0.81} & 0.08  & 0.35  & 0.15  & 0.94  & 0.19   \\
     & Pinch and Twirl & {\color{orange} 0.73 }  & 0.06  & 0.15  & 0.06  & 0.51  & 0.00  \\
     & Caustic Refraction & {\color{orange} 0.88 } & 0.09  & 0.18  & 0.07  & 0.48  & 0.09  \\
     & Fish Eye & {\color{orange} 0.68 }  & 0.08  & 0.25  & 0.10  & 0.88  & 0.05  \\
     & Water Drop & {\color{orange} 0.89 } & 0.10  & 0.28  & 0.13  & 0.97  & 0.03  \\
     & Ripple & {\color{orange} 0.89 } & 0.12  & 0.40  & 0.18  & 1.08  & 0.07  \\
     \midrule 
   Digital 
   & Color Balance & 0.18 & 0.09  & 0.74  & 0.45  & 0.20  & 0.29  \\
    & Quadrilateral No Bars & {\color{orange} 0.77 } & 0.13  & 0.57  & 0.26  & 1.62  & 0.02  \\
    & Perspective No Bars & {\color{orange} 0.75 } & 0.16  & 0.80  & 0.30  & {\color{orange} 2.08 } & 0.03  \\
    & Bleach Bypass & 0.25 & 0.08  & 0.33  & 0.13  & 0.94  & 0.07  \\
    & Technicolor & {\color{red} 0.90} & {\color{orange} 0.21 }  & 0.75  & 0.45  & 0.94  & 0.24  \\
    & Pseudocolor & 0.21 & 0.11  & 0.93  & 0.53  & 0.94  & {\color{orange} 0.41}  \\
    & Hue Shift & {\color{orange} 0.55 } & {\color{orange} 0.26 } & 1.53  & {\color{orange} 0.92 } & 0.94  & {\color{orange} 0.36}  \\
    & Color Dither & 0.17 & 0.06  & 0.02  & 0.00  & 0.00  & 0.04  \\
    \midrule
    Occlusion 
    & Single Frequency GreyScale & 0.18 & 0.12  & 0.04  & 0.02  & 0.03  & 0.14   \\
    & Checkerboard Cutout & 0.42 & 0.07  & 0.24  & 0.14  & 0.08  & 0.05   \\
    & Sparkles & {\color{orange} 0.77 } & 0.14  & 0.77  & 0.30  & 0.59  & 0.13  \\
    & Inverse Sparkles & {\color{orange} 0.72 } & {\color{orange} 0.26 }  & {\color{orange} 1.75 }  & 0.61  &  0.50  & {\color{orange} 0.33}  \\
    & Lines & 0.43  & 0.10  & 0.65  & 0.35  & 0.67  & 0.20  \\
    \bottomrule
    \end{tabular}
    \caption{Threats evaluated on Imagenet-$\bar{C}$ corruptions in \Cref{fig:lionfish_corruptions_c_bar}.}
    \label{tab:LionFish_c_bar}
\end{table}

\newpage 

\section{Comparison of Average Threat Statistics}
\begin{figure}[h]
    \centering
    \begin{minipage}{0.4\linewidth}
        \centering     
        \includegraphics[width=\textwidth, trim={0cm 0cm 7cm 0cm},clip]{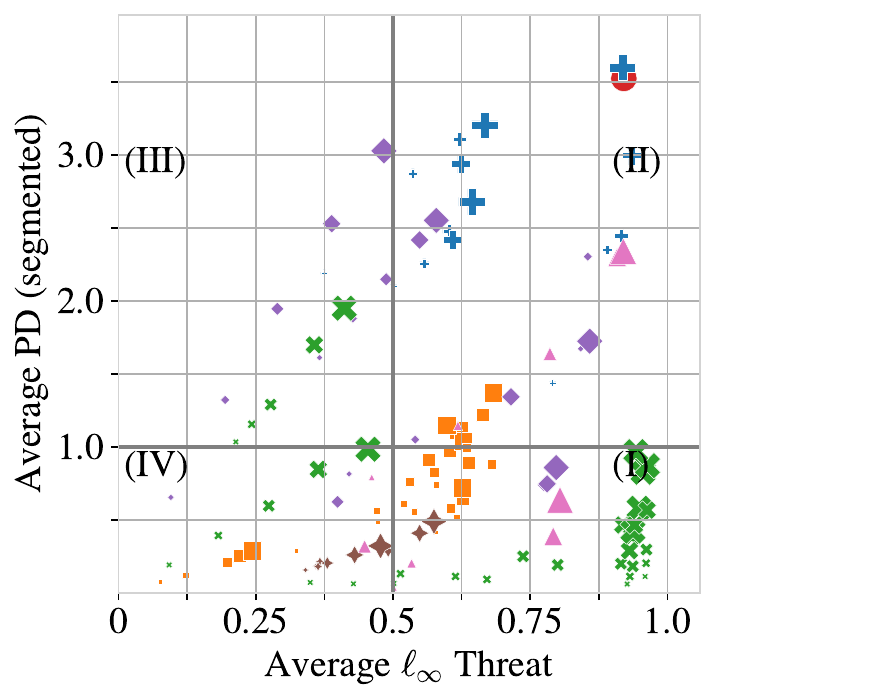}
        \subcaption{PD-S vs $\ell_\infty$}
        \label{fig:PD-S-vs-Linf}
    \end{minipage}
    \hspace{-1cm} 
    \begin{minipage}{0.15\linewidth}
        \centering
        \includegraphics[width=\textwidth, trim={0.1cm 0cm 0.1cm 0cm},clip]{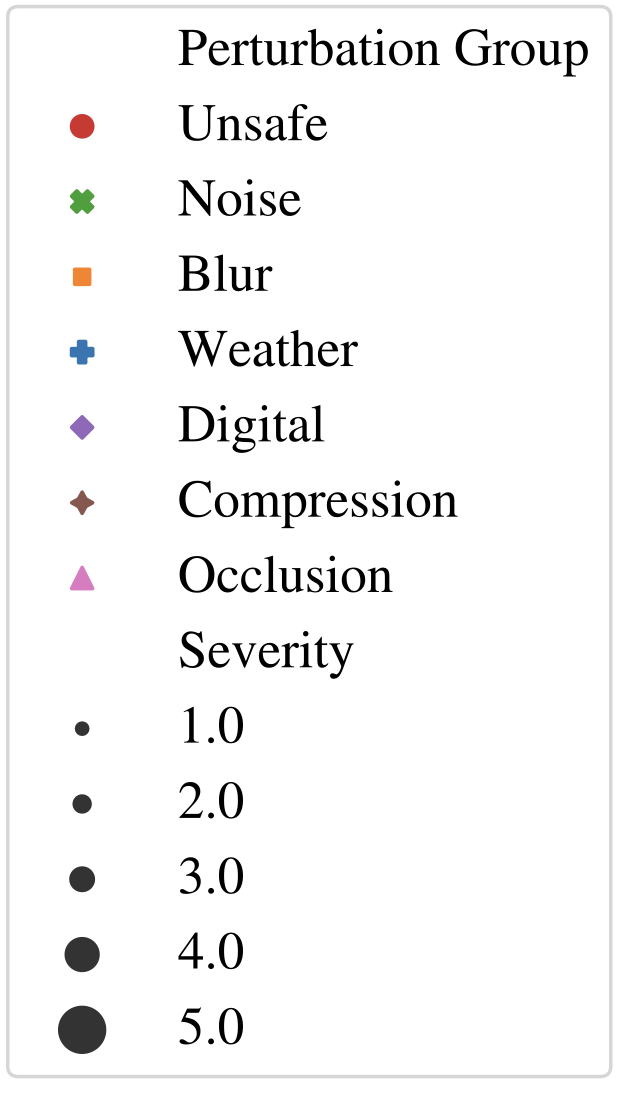}
    \end{minipage}
    \begin{minipage}{0.4\linewidth}
        \centering     
        \includegraphics[width=\textwidth, trim={0cm 0cm 7cm 0cm},clip]{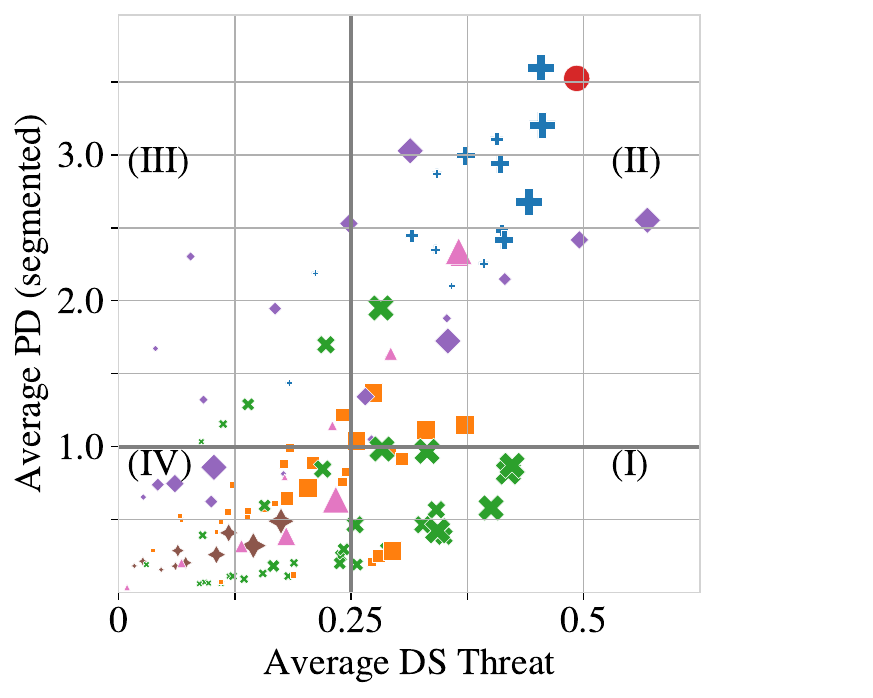}
        \subcaption{PD-S vs DS}
        \label{fig:PD-S-vs-DS}
    \end{minipage}
    \caption{Comparison of PD-S threat.}
    \label{fig:comparison-PD-S}
\end{figure}

\begin{figure}[h]
    \centering
    \begin{minipage}{0.4\linewidth}
        \centering     
        \includegraphics[width=\textwidth, trim={0cm 0cm 7cm 0cm},clip]{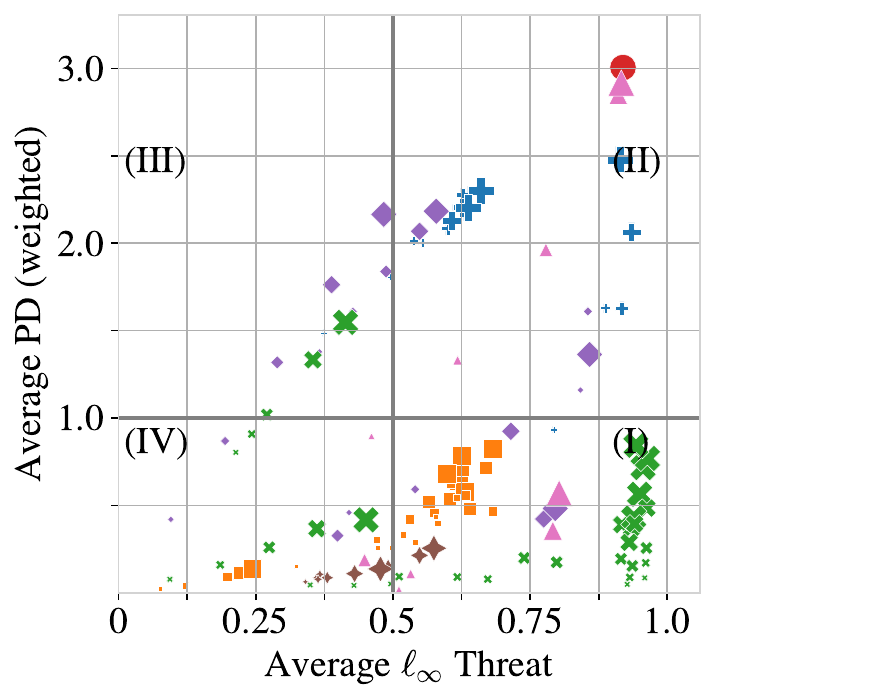}
        \subcaption{PD-W vs $\ell_\infty$}
        \label{fig:PD-W-vs-Linf}
    \end{minipage}
    \hspace{-1cm} 
    \begin{minipage}{0.15\linewidth}
        \centering
        \includegraphics[width=\textwidth, trim={0.1cm 0cm 0.1cm 0cm},clip]{Figures/legend-appendix.png}
    \end{minipage}
    \begin{minipage}{0.4\linewidth}
        \centering     
        \includegraphics[width=\textwidth, trim={0cm 0cm 7cm 0cm},clip]{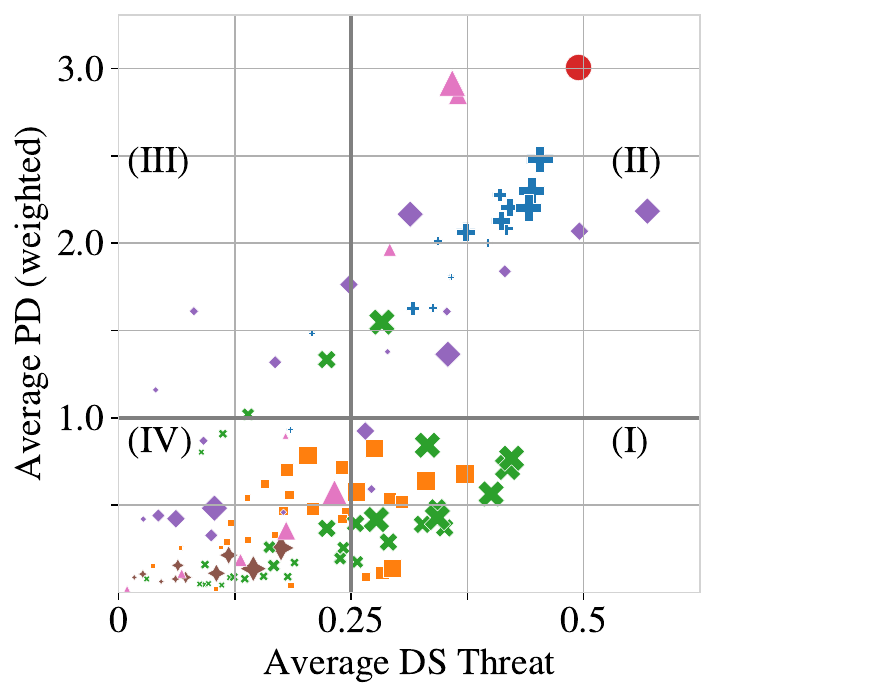}
        \subcaption{PD-W vs DS}
        \label{fig:PD-W-vs-DS}
    \end{minipage}
    \caption{Comparison of PD-W threat.}
    \label{fig:comparison-PD-W}
\end{figure}

\begin{figure}[h]
    \begin{minipage}{0.45\linewidth}
    \flushleft
        \includegraphics[width=\textwidth,trim={2.2cm 1cm 3cm 0.2cm},clip]{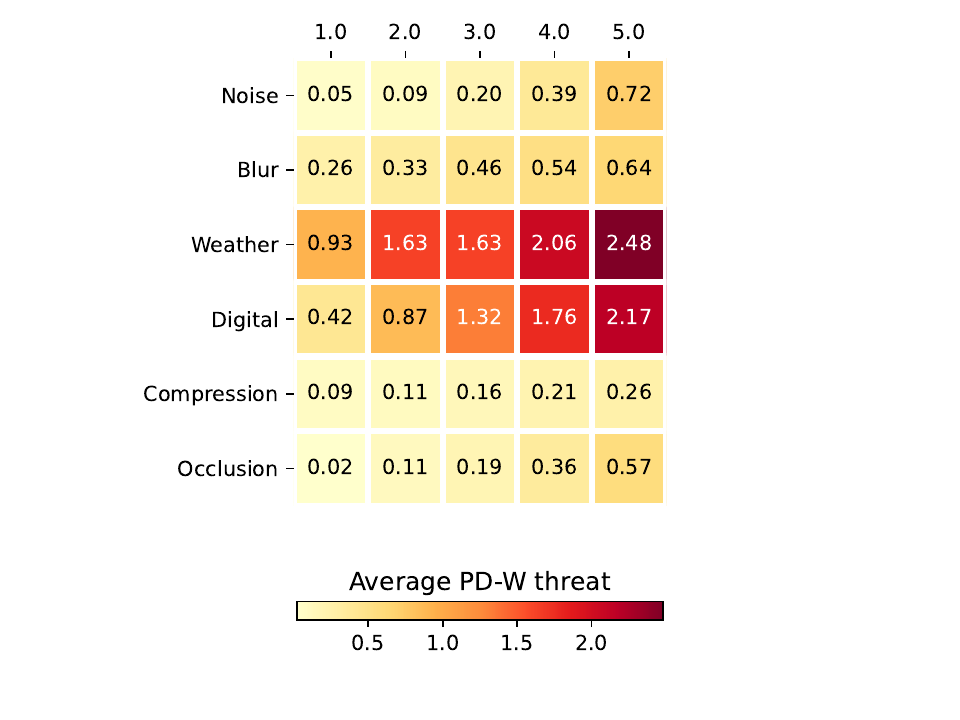}
        \label{fig:severity-PD-W}
    \end{minipage}%
    \begin{minipage}{0.45\linewidth}
    \centering
        \includegraphics[width=\textwidth,trim={2.2cm 1cm 3cm 0.2cm},clip]{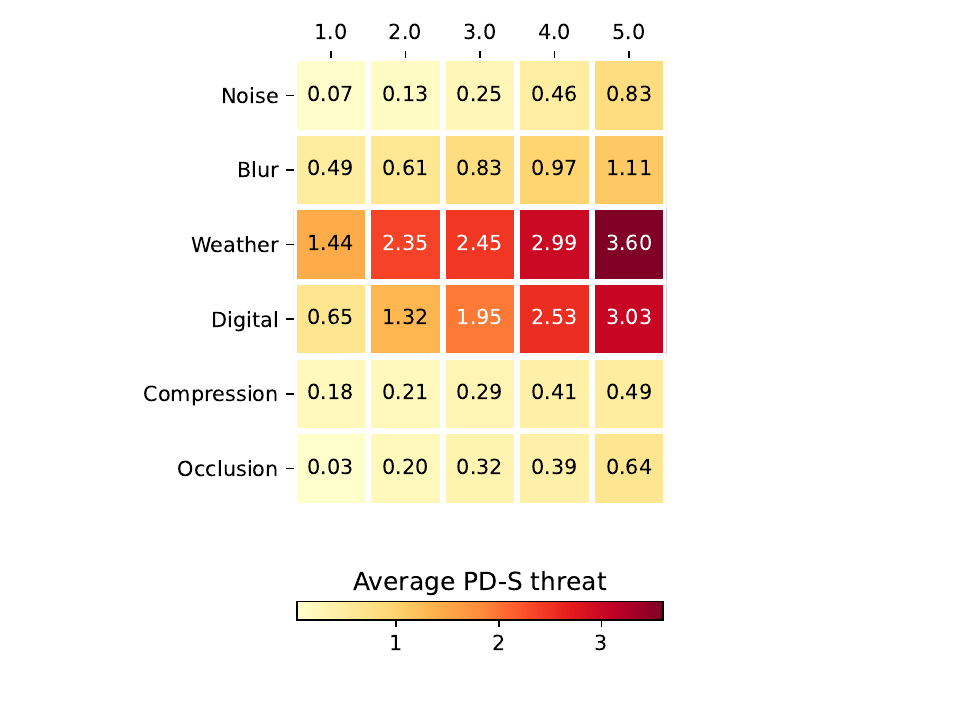}
        \label{fig:severity-PD=S}
    \end{minipage}%
    \caption{Heatmap of threat models vs severity levels of corruption groups.}
    \label{fig:severity-comparison-PD-2}
\end{figure}

\end{document}